\pdfoutput=1
\documentclass[10pt,twocolumn,letterpaper]{article}

\usepackage[pagenumbers]{cvpr}
\usepackage{amsthm}
\usepackage{mathtools}
\usepackage{tabularx}
\usepackage{microtype}
\usepackage{stfloats}

\definecolor{bpOurs}{RGB}{0,114,178}
\definecolor{bpPrior}{RGB}{213,94,0}
\definecolor{bpData}{RGB}{0,158,115}
\definecolor{bpAccent}{RGB}{230,159,0}
\definecolor{bpGray}{RGB}{110,110,110}

\newtheoremstyle{cvprthm}{2.0pt}{2.0pt}{\itshape}{}{\bfseries}{.}{0.4em}{}
\theoremstyle{cvprthm}
\newtheorem{theorem}{Theorem}
\newtheorem{lemma}{Lemma}
\newtheorem{proposition}{Proposition}
\newtheorem{corollary}{Corollary}

\makeatletter
\renewenvironment{proof}[1][\proofname]{\par
  \pushQED{\qed}%
  \normalfont \topsep1.5\p@\@plus0.8\p@\relax
  \trivlist
  \item[\hskip\labelsep\itshape #1\@addpunct{.}]\ignorespaces
}{%
  \popQED\endtrivlist\@endpefalse
}
\makeatother

\AtBeginDocument{%
  \captionsetup[subtable]{aboveskip=2pt,belowskip=2pt}%
  \setlength{\abovedisplayskip}{4.5pt plus 1.5pt minus 1.5pt}%
  \setlength{\belowdisplayskip}{4.5pt plus 1.5pt minus 1.5pt}%
  \setlength{\abovedisplayshortskip}{3pt plus 1pt}%
  \setlength{\belowdisplayshortskip}{4pt plus 1pt}%
}

\renewcommand{\paragraph}[1]{\vspace{.35em}\noindent\textbf{#1}}
\definecolor{cvprblue}{rgb}{0.21,0.49,0.74}
\usepackage[pagebackref,breaklinks,colorlinks,allcolors=cvprblue]{hyperref}
\hypersetup{
  pdftitle={AdapToC: A Unified Framework for Multi-Task Visual Token Communication},
  pdfauthor={Zhicheng Liu, Chenliang Wu, Zhouxiang Zhao, Zhaohui Yang, Zhaoyang Zhang},
}
\usepackage{algorithm}
\usepackage{algpseudocode}
\usepackage{placeins}
\makeatletter
\renewcommand{\theHALG@line}{\thealgorithm.\arabic{ALG@line}}
\makeatother

\title{AdapToC: A Unified Framework for Multi-Task Visual Token Communication}

\author{Zhicheng Liu \quad Chenliang Wu \quad Zhouxiang Zhao \quad Zhaohui Yang \quad Zhaoyang Zhang\\
Zhejiang University
}

\begin{document}
\twocolumn[{
  \renewcommand\twocolumn[1][]{#1}
  \maketitle
  \begin{center}
  \centering
  \vspace{-0.8em}
  \includegraphics[width=0.98\linewidth]{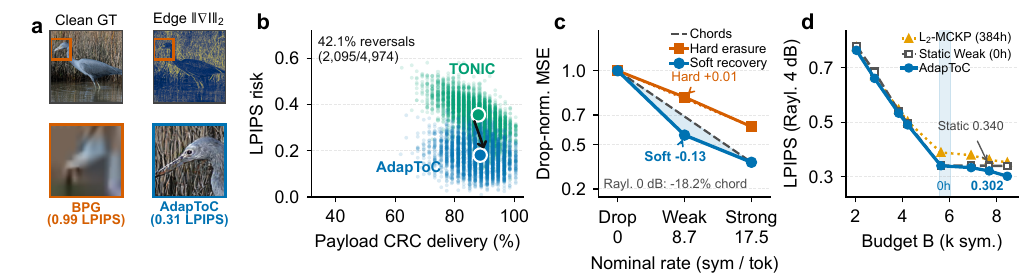}
  \captionsetup{hypcap=false}
  \captionof{figure}{\textbf{Perceptual quality over wireless fading channels depends on soft token recovery across budget regimes.} (\textbf{a}) At $6\,\text{dB}$ Rayleigh, BPG ($0.99$ LPIPS) blurs contours while AdapToC ($0.31$) preserves detail. (\textbf{b}) Higher CRC delivery often reverses paired LPIPS ($42.1\%$ across $4{,}974$ images). (\textbf{c}) Hard erasure lifts \textsc{Weak} above the chord; soft recovery pulls it below ($-18.2\%$ at $0\,\text{dB}$). (\textbf{d}) Across same-receiver Rayleigh budgets ($4\,\text{dB}$, $N{=}512$), AdapToC beats static \textsc{Weak} and $L_2$-MCKP in severe deficit ($B{\le}4{,}224$; header truncation leads at $5{,}184$), matches $0\text{h}$ inside $[B_{\text{full}},B_{\text{up}})$, and avoids saturation above $B_{\text{up}}$ ($0.302$ vs.\ $0.340$).}
  \label{fig:teaser}
  \vspace{0.3em}
\end{center}
}]

\begin{abstract}
Pretrained discrete visual tokenizers supply compact latents for synthesis, recognition, and embodied perception, yet wireless token streaming remains fragile. Under fading and noisy pilots, hard cyclic redundancy check (CRC) erasures turn corrupted blocks into missing tokens while frozen decoders amplify edge errors, which can collapse unequal error protection (UEP) into a binary choice between strong coding and omission under severe fading. Soft bit posteriors break this collapse. We present \textbf{AdapToC}, a single-pass transceiver that restores multi-tier rate-distortion tradeoffs without altering tokenizer weights or differentiating through pixel decoders. We align codebook geometry to the channel via offline frequency-weighted Pseudo-Gray relabeling and Halton dispersion to contract bit and burst errors with zero overhead. Because soft recovery pulls intermediate rates onto the three-tier convex hull, our unit-gated AdaLN sender schedules protection tiers across channel budgets, while our pilot-calibrated receiver feeds factorized codebook posteriors into a ConvNeXt-attention restorer with CRC bypass. We \textbf{co-train AdapToC solely in latent space} via counterfactual utility estimation and codebook cross-entropy regularization. On identical VQGAN-$f_{16}$ tokens, AdapToC outperforms TONIC and adaptive TokenCom+AMC, cuts multi-SNR Rayleigh LPIPS from $0.6551$ to $0.3312$ within the frozen-tokenizer PSNR ceiling, and saves $43.8\%$ Rician-$K6$ symbols at matched quality. Our single-pass restorer runs $8.3\times$ faster than iterative resampling, raises accuracy while accelerating decoder-free recognition $20.6\times$, and improves fidelity plug-and-play across TiTok, VAR, and zero-shot NuPlan.
\end{abstract}

\section{Introduction}
\label{sec:intro}

Discrete visual tokenizers (VQGAN~\cite{esser2021vqgan}, VAR~\cite{tian2024var}, TiTok~\cite{yu2024titok}) encode images into codebook indices for synthesis, recognition, and world modeling~\cite{drivingworld2024}. Over fading orthogonal frequency-division multiplexing (OFDM) links~\cite{liu2026tonic,qiao2025tokencom}, packet delivery statistics fail to predict human perception since higher cyclic redundancy check (CRC) delivery often reverses paired LPIPS~\cite{zhang2018lpips} rankings (Fig.~\ref{fig:teaser}b).

\begin{figure*}[t]
  \centering
  \setlength{\abovecaptionskip}{2pt}
  \setlength{\belowcaptionskip}{1pt}
  \includegraphics[width=0.95\linewidth]{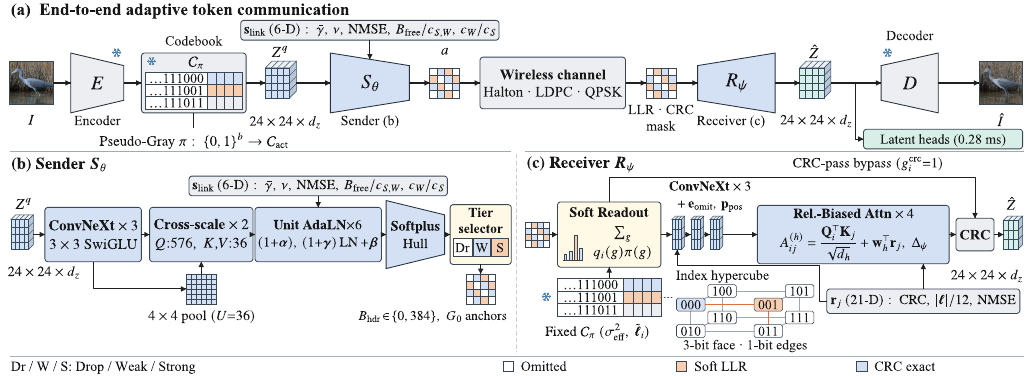}
  \caption{\textbf{AdapToC combines bit-aligned token transport, adaptive tier selection, and single-pass soft recovery.} (\textbf{a}) System path across $\pi$, $S_\theta$, $R_\psi$, and frozen $D$ or task heads; (\textbf{b}) unit-gated AdaLN tier scheduling and hull pooling; (\textbf{c}) soft readout with ConvNeXt refinement, reliability attention, and CRC bypass.}
  \label{fig:overview}
\end{figure*}

Two coupled bottlenecks cause this degradation (Fig.~\ref{fig:teaser}a,b), as frozen decoders amplify high-frequency perturbations over smooth errors~\cite{yao2025vavae} while pixel backpropagation is costly and Euclidean mean squared error (MSE) blurs contours. On the channel side, pilot noise inflates bit log-likelihood ratios (LLRs), and hard CRC gating turns failed blocks into missing tokens. Near moderate coding thresholds, intermediate protection incurs higher distortion than interpolating between omitted (\textsc{Drop}) and strongly coded (\textsc{Strong}) tokens and collapses unequal error protection (UEP) to a binary choice (Fig.~\ref{fig:teaser}c).

Soft bit recovery breaks this collapse because retaining calibrated bit posteriors on CRC-failed blocks keeps errors local, pulls intermediate tiers onto the lower convex hull (Fig.~\ref{fig:teaser}c), and supports adaptive protection across budgets (Fig.~\ref{fig:teaser}d). We instantiate this in \textbf{AdapToC}, a single-pass transceiver co-trained in latent space with frozen tokenizer weights (Fig.~\ref{fig:overview}). Over the training-fitted active codebook ($K{=}2^b{=}1{,}024$), we apply Pseudo-Gray matching to cut single-bit MSE by $61.1\%$, Halton ordering to disperse bursts, and prefix framing to pack extra tokens per block. Across link states and budgets, our unit-gated adaptive layer norm (AdaLN) sender schedules tiers, while our pilot-calibrated receiver feeds factorized codebook posteriors into a ConvNeXt-attention restorer with CRC bypass.

Unlike continuous joint source-channel coding (JSCC)~\cite{tung2022deepjsccq,wu2025mambajscc}, AdapToC preserves frozen discrete tokenizers and outperforms separated codecs, static UEP~\cite{liu2026tonic}, and iterative TokenCom~\cite{qiao2025tokencom} in one pass (Tab.~\ref{tab:main}). Across $4{,}974$ ImageNet images on shared VQGAN-$f_{16}$ tokens (zero-header $B=5{,}632$), it cuts multi-SNR (signal-to-noise ratio) Rayleigh LPIPS over static UEP and adaptive modulation/coding (TokenCom+AMC) from $0.6346$/$0.6551$ to $0.3312$, lowers $8\,\text{dB}$ Rayleigh rFID over $3.5\times$, and saves $43.8\%$ to $61.7\%$ Rician-$K6$ symbols. Our restorer runs $8.3\times$ faster than TokenCom, raises fading accuracy while speeding recognition $20.6\times$, and improves fidelity plug-and-play (PnP) across TiTok, VAR, and zero-shot NuPlan (Tab.~\ref{tab:generality}).

Our contributions span three areas:
\begin{itemize}
  \item \textbf{Soft-recovery rate-distortion conditions} connect decoder perceptual sensitivity to soft posteriors, establish when intermediate tiers enter the lower convex hull, and characterize header-limited granularity (Fig.~\ref{fig:teaser}c,d; Sec.~\ref{sec:theory}).
  \item \textbf{Bit-aligned token transceiver} unifies Pseudo-Gray labeling, prefix packing, Halton dispersion, AdaLN tier scheduling, and factorized $K$-way readout with a ConvNeXt-attention restorer (Secs.~\ref{sec:geometry} to~\ref{sec:receiver}).
  \item \textbf{Counterfactual latent co-training} pairs uniform-rollout utility estimation with codebook cross-entropy regularization to align tier scheduling and restoration without decoder gradients (Secs.~\ref{sec:cotraining}, \ref{sec:exp}).
\end{itemize}
\section{Related Work}
\label{sec:related}

\subsection{Visual Tokenizers and Latent Restoration}
Frozen discrete tokenizers~\cite{esser2021vqgan,tian2024var,yu2024titok,chen2025softvq,yao2025vavae} encode latents for synthesis, recognition, and embodied simulation~\cite{drivingworld2024}. Prior restorers avoid decoder tuning: VarFormer~\cite{wang2025varformer} aligns VAR hierarchies, SILO~\cite{raphaeli2025silo} learns degradation operators~\cite{zhang2025daps}, DicFace~\cite{chen2025dicface} blends codebooks, and APS~\cite{rout2026aps} guides remasking from anchors. Unlike iterative binary-erasure restorers, AdapToC resolves CRC anchors, soft LLRs, and omitted blocks in one pass.

\subsection{Loss-Resilient Neural Image Coding}
Learned codecs trade rate, distortion, and perception~\cite{he2022elic,jia2024glc,cao2026progic,blau2019rdp}, while RDVQ~\cite{jiang2026rdvq} separates soft rate relaxation from hard quantization. Packet-resilient coders~\cite{sha2025lossresilientlic,wei2026dispersing} scatter latents across packets using erasure flags that discard bit reliability. Unlike channel-optimized vector quantization~\cite{farvardin1990covq}, which shifts centroids, AdapToC freezes codebooks under low-density parity-check (LDPC) OFDM packets via Pseudo-Gray, prefix packing, and Halton dispersion.

\subsection{Wireless Visual Communication}
Continuous JSCC~\cite{yang2025swinjscc,wu2025mambajscc} and DeepJSCC-Q~\cite{tung2022deepjsccq} avoid digital cliffs~\cite{bellard2014bpg} via pixel encoders instead of frozen tokenizers~\cite{xie2026videorae}. Token transport uses static UEP with CRC erasures (TONIC~\cite{liu2026tonic}), iterative resampling (TokenCom~\cite{qiao2025tokencom,chang2022maskgit}), autoregressive coding (Ada-TokenCom~\cite{zhang2026adatokencom}), cross-layer forward error correction (FEC)~\cite{zhang2026imageclsec}, and counterfactual ordering~\cite{guo2026gcrc}. Instead of multi-pass decoding, AdapToC restores soft posteriors across adaptive tiers in one pass.
\section{Problem Formulation and Theory}
\label{sec:theory}

Building on the packet-erasure and distortion--perception limitations identified above, we formalize discrete token transmission over fading orthogonal frequency-division multiplexing (OFDM) channels (Sec.~\ref{sec:sysmodel}), connect decoder perceptual risk to conditional posterior moments (Lemma~\ref{lem:anisotropy}), and derive exact convex-hull and header-overhead conditions for budget-adaptive tier selection (Theorem~\ref{thm:unfolding}, Corollary~\ref{cor:granularity}).

\subsection{Wireless Discrete Token Channel}
\label{sec:sysmodel}
A frozen tokenizer maps an image $I \in \mathbb{R}^{H\times W\times 3}$ via encoder $E(I)$ to $L$ latent embeddings $Z^q = [z_1^q,\dots,z_L^q]^\top \in \mathbb{R}^{L\times d_z}$ supported on an active $b$-bit codebook $\mathcal{C}_{\text{act}} = \{v_0,\dots,v_{K-1}\}$ ($K = 2^b$, nearest-neighbor projected from the training codebook $\mathcal{C}$ with $0\%$ unseen codes across all evaluation splits) and reconstructed by frozen pixel decoder $D$. Under bijection $\pi:\{0,1\}^b \to \mathcal{C}_{\text{act}}$, token $i$ carries bit codeword $g_i^\star = \pi^{-1}(z_i^q)$. Sparse anchors $G_0$ occupy a coarse spatial lattice, while the remaining tokens partition into $U$ spatial units assigned protection tiers $a_u \in \mathcal{A} = \{a_0,\dots,a_M\}$ with symbol costs $0 = c_0 < c_1 < \dots < c_M$ ($a_0 = \textsc{Drop}$) under total frame budget $B_{\text{tot}}$ ($B$) and channel interleaving map $s(i)$.

On subband $s$, the received complex baseband symbol satisfies $y_s = h_s x_s + n_s$ with $\mathbb{E}[|x_s|^2] = 1$, $\mathbb{E}[|h_s|^2] = 1$, thermal noise $n_s \sim \mathcal{CN}(0,\sigma_n^2)$ ($\sigma_n^2 > 0$), fading severity $\nu_{\text{fad}} \in [0,1]$ ($0$ for additive white Gaussian noise (AWGN), $1$ for Rayleigh fading), and average signal-to-noise ratio (SNR) $\bar{\gamma}$ (dB). Given orthogonal linear minimum MSE (LMMSE) channel state information (CSI) $\hat{h}_s = h_s - e_s$ with subband error variance $\sigma_{e,s}^2$ (and link-level average pilot $\text{NMSE} = \mathbb{E}[|e_s|^2]$), the post-estimation effective SNR is $\hat{\gamma}_s = |\hat{h}_s|^2/(\sigma_n^2+\sigma_{e,s}^2)$ ($\hat{\gamma}_{s,\text{dB}} = 10\log_{10}\max(\hat{\gamma}_s, \gamma_{\min})$ with finite floor $\gamma_{\min} > 0$). Low-density parity-check (LDPC) decoding outputs a cyclic redundancy check (CRC) flag $g_i^{\text{crc}} \in \{0,1\}$, transmission indicator $g_i^{\text{sent}} \in \{0,1\}$ (with $g_i^{\text{crc}} \le g_i^{\text{sent}}$), hard bits $\hat{g}_i \in \{0,1\}^b$, and bit log-likelihood ratios (LLRs) $\boldsymbol{\ell}_i \in \mathbb{R}^b$, partitioning tokens into verified ($g_i^{\text{crc}} = 1$), soft-uncertain ($g_i^{\text{crc}} = 0, g_i^{\text{sent}} = 1$), and omitted ($g_i^{\text{sent}} = 0$) states.

\subsection{Perceptual Risk and Soft Posterior Bounds}
\label{sec:anisotropy}
Given channel observations $O$ and restored latents $\hat{Z} = [\hat{z}_1,\dots,\hat{z}_L]^\top \in \mathbb{R}^{L\times d_z}$, let $F(z) = \Phi(D(\operatorname{unvec}(z))) \in \mathbb{R}^{d_F}$ denote the perceptual feature map~\cite{zhang2018lpips} (incorporating per-layer normalization, channel weighting, and spatial averaging) evaluated on vectorized latents $z = \operatorname{vec}(Z^q) \in \mathbb{R}^{Ld_z}$ and $\hat{z}(O) = \operatorname{vec}(\hat{Z})$, with posterior mean $\mu = \mu(O) = \mathbb{E}[z \mid O]$ and covariance $\Sigma = \Sigma(O) = \operatorname{Cov}(z \mid O)$. Lemma~\ref{lem:anisotropy} establishes that the local quadratic surrogate of perceptual risk is minimized by $\mu(O)$, while nearest-codeword projection preserves strict multiplicative MSE stability whereas maximum a posteriori (MAP) decoding admits no codebook-size-independent bound.

\begin{lemma}[Local Perceptual Risk and Global MSE Bounds]
\label{lem:anisotropy}
Assume $F$ is $C^2$ on a convex neighborhood $\mathcal{B}_r(\mu(O))$ containing $\hat{z}(O)$ and containing $z$ almost surely given $O$, with bounded Jacobian operator norm $\|J_F(\mu)\|_{\textup{op}} \le L_J < \infty$ and bounded quadratic Hessian norm $\sup_{\xi \in \mathcal{B}_r(\mu),\, \|u\|_2 = 1} \|\nabla^2 F(\xi)[u,u]\|_2 \le H_F < \infty$, and finite conditional fourth moment $\mathbb{E}[\|z-\mu\|_2^4 \mid O] < \infty$ (automatic on finite $\mathcal{C}_{\textup{act}}^L$). With $M_F = J_F(\mu)^\top J_F(\mu) \succeq 0$ and $\|v\|_{M_F}^2 \triangleq v^\top M_F v$, expanding $F$ around $\mu(O)$ yields
\begin{equation}
  \mathbb{E}[\|F(z)\!-\!F(\hat{z})\|_2^2 \mid O]
  = \|\hat{z}\!-\!\mu\|_{M_F}^2 \!+\! \operatorname{tr}(M_F\Sigma(O)) + R_3(O),
  \label{eq:anisotropy}
\end{equation}
with $|R_3(O)| \le (2L_J H_F + \tfrac{1}{2}r H_F^2)(\mathbb{E}[\|z-\mu\|_2^3 \mid O] + \|\hat{z}-\mu\|_2^3)$. The quadratic surrogate $\|\hat{z}-\mu\|_{M_F}^2 + \operatorname{tr}(M_F\Sigma(O))$ is minimized at $\hat{z}^\star = \mu(O)$ (with minimizer set $(\mu(O) + \ker J_F(\mu)) \cap \mathcal{B}_r(\mu(O))$ if $J_F(\mu)$ is rank-deficient). Globally (without locality), for any $M \succ 0$, $\mathbb{E}[\|z-\hat{z}\|_M^2 \mid O]$ has unique minimizer $\hat{z}^\star = \mu(O)$. Euclidean nearest-codeword projection $\Pi(\mu(O)) \in \mathcal{C}_{\textup{act}}^L$ increases expected MSE at most $2\times$ over $\mu(O)$ ($4\times$ sample-wise), whereas non-Euclidean MAP decoding admits no codebook-size-independent finite multiplicative bound (and is unbounded sample-wise even on a fixed codebook).
\end{lemma}

\begin{proof}
\emph{Local perceptual quadratic pullback.} Applying Taylor's theorem with integral remainder around $\mu$ gives $F(z) - F(\hat{z}) = J_F(\mu)[(z-\mu) + (\mu-\hat{z})] + (\boldsymbol{\zeta}_z - \boldsymbol{\zeta}_{\hat{z}})$, where $\boldsymbol{\zeta}_v = \int_0^1 (1-t)\nabla^2 F(\mu + t(v-\mu))[v-\mu,v-\mu]\,\mathrm{d}t$ satisfies $\|\boldsymbol{\zeta}_v\|_2 \le \tfrac{1}{2} H_F \|v-\mu\|_2^2$ for $v \in \{z,\hat{z}\}$. Taking squared norms and conditioning on $O$, the linear cross-term vanishes because $\mathbb{E}[z-\mu \mid O] = 0$, leaving Eq.~\eqref{eq:anisotropy} with remainder $R_3(O) = 2\mathbb{E}[\langle J_F(\mu)(z-\hat{z}),\,\boldsymbol{\zeta}_z-\boldsymbol{\zeta}_{\hat{z}}\rangle \mid O] + \mathbb{E}[\|\boldsymbol{\zeta}_z-\boldsymbol{\zeta}_{\hat{z}}\|_2^2 \mid O]$. Setting $r_z \triangleq \|z-\mu\|_2$ and $r_{\hat{z}} \triangleq \|\hat{z}-\mu\|_2$, Cauchy--Schwarz, the triangle inequality $\|z-\hat{z}\|_2 \le r_z+r_{\hat{z}}$, the elementary bound $(r_z-r_{\hat{z}})^2(r_z+r_{\hat{z}}) \ge 0 \implies r_z^2 r_{\hat{z}} + r_z r_{\hat{z}}^2 \le r_z^3 + r_{\hat{z}}^3$, and $z,\hat{z} \in \mathcal{B}_r(\mu)$ ($r_v^4 \le r r_v^3$) yield $|R_3(O)| \le (2L_J H_F + \tfrac{1}{2}r H_F^2)(\mathbb{E}[r_z^3 \mid O] + r_{\hat{z}}^3)$; finite fourth moments ensure finite cubic and quartic remainder moments via H\"older's inequality. Since $M_F \succeq 0$, the quadratic surrogate $\|\hat{z}-\mu\|_{M_F}^2 + \operatorname{tr}(M_F\Sigma)$ is minimized precisely when $\hat{z}-\mu \in \ker(J_F(\mu))$, including $\hat{z}^\star = \mu(O)$ (note that the exact nonlinear risk with $R_3(O)$ can shift its minimizer away from $\mu(O)$ when $H_F > 0$).

\emph{Global optimality and sharp $2\times$/$4\times$ Euclidean projection bounds.} Without any locality assumption on $F$, writing $z-\hat{z} = (z-\mu) + (\mu-\hat{z})$ for any $M \succ 0$ eliminates the cross-term via $\mathbb{E}[(z-\mu)^\top M(\mu-\hat{z}) \mid O] = 0$ and gives the exact bias--variance identity $\mathbb{E}[\|z-\hat{z}\|_M^2 \mid O] = \mathbb{E}[\|z-\mu\|_M^2 \mid O] + \|\hat{z}-\mu\|_M^2$, uniquely minimized at $\hat{z}^\star = \mu(O)$. Exploiting the discrete support $z \in \mathcal{C}_{\text{act}}^L$, nearest-codeword optimality of $\Pi(\mu) \in \arg\min_{v \in \mathcal{C}_{\text{act}}^L}\|v-\mu\|_2$ ensures $\|\mu - \Pi(\mu)\|_2 \le \|z-\mu\|_2$ almost surely. By the triangle inequality, $\|z-\Pi(\mu)\|_2 \le \|z-\mu\|_2 + \|\mu-\Pi(\mu)\|_2 \le 2\|z-\mu\|_2$, yielding $\|z-\Pi(\mu)\|_2^2 \le 4\|z-\mu\|_2^2$ sample-wise, while setting $M = I$ in the bias--variance identity with $\|\mu-\Pi(\mu)\|_2^2 \le \mathbb{E}[\|z-\mu\|_2^2 \mid O]$ tightens the expected bound to $\mathbb{E}[\|z-\Pi(\mu)\|_2^2 \mid O] \le 2\mathbb{E}[\|z-\mu\|_2^2 \mid O]$. Both constants are sharp: on $\mathcal{C}_{\text{act}}^L = \{+v,-v\}$ ($v \neq 0$) with equiprobable antipodal posterior $P(z=+v \mid O) = P(z=-v \mid O) = 1/2$, we have $\mu = 0$ and tie-broken $\Pi(0) = +v$, giving $2\|v\|_2^2$ expected MSE ($2\times$) and $4\|v\|_2^2$ when $z=-v$ ($4\times$).

\emph{Codebook-dependent MAP growth and sample-wise unboundedness.} On a fixed finite codebook of size $K^L$, writing $p = P(z = z_{\text{MAP}} \mid O) \ge K^{-L}$ and applying Jensen's inequality with the bias--variance identity gives $\text{MSE}_{\text{MAP}} \le \text{MMSE}/p \le K^L\,\text{MMSE}$ (or $K\,\text{MMSE}$ under additive per-token MAP), so MAP is bounded by the support cardinality but admits no codebook-size-independent constant. Indeed, for any integer $N_c \ge 2$ (e.g., $N_c = 2^b - 1$), scale $A > 0$, and $0 < \delta < N_c/(N_c+1)$, consider a scalar posterior over $N_c+1$ codewords $\{0\} \cup \{A+j\varepsilon\}_{j=0}^{N_c-1}$ ($\varepsilon > 0$) with $P(z=0 \mid O) = p_0 \triangleq 1/(N_c+1) + \delta$ and $P(z=A+j\varepsilon \mid O) = (1-p_0)/N_c$ for $j \in \{0,\dots,N_c-1\}$. Because $p_0 > (1-p_0)/N_c$, MAP uniquely selects $\hat{z}_{\text{MAP}} = 0$ even though mass $1-p_0$ concentrates near $A$. As $\varepsilon \to 0^+$, $\text{MSE}_{\text{MAP}} \to (1-p_0)A^2$ whereas $\mu \to (1-p_0)A$ with $\text{MMSE} \to p_0(1-p_0)A^2$, yielding
\[
  \frac{\text{MSE}_{\text{MAP}}}{\text{MMSE}}
  \;\xrightarrow{\;\varepsilon \to 0^+\;}\; \frac{1}{p_0}
  \;\xrightarrow{\;\delta \to 0^+\;}\; N_c + 1
  \;\xrightarrow{\;N_c \to \infty\;}\; \infty,
\]
already exceeding $2\times$ at $N_c = 2$ ($1/p_0 \to 3$). Moreover, even on a fixed $3$-codeword support $\{0,1,3\}$ with probabilities $(1/2, 1/4, 1/4)$, $\mu = 1$ and $z_{\text{MAP}} = 0$, so sample-wise error ratio at $z = 1$ is unbounded.
\end{proof}

\begin{figure*}[t]
  \centering
  \setlength{\abovecaptionskip}{2pt}
  \setlength{\belowcaptionskip}{1pt}
  \includegraphics[width=0.96\linewidth]{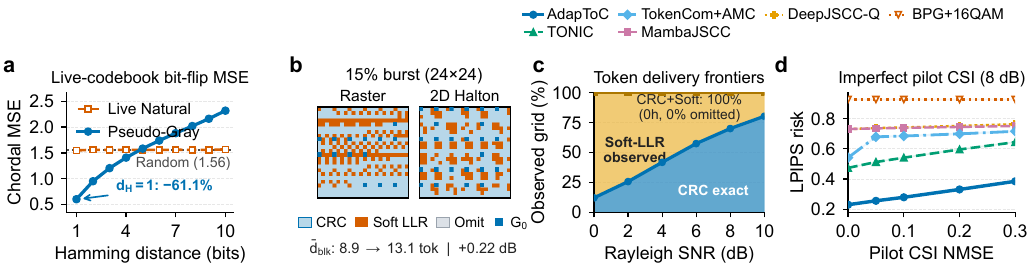}
  \caption{\textbf{Diagnostics verify codebook alignment, burst dispersion, soft coverage, and CSI tolerance.} (\textbf{a}) Pseudo-Gray cuts $1{,}024$-code $d_H{=}1$ MSE by $61.1\%$. (\textbf{b}) Halton widens $15\%$ burst spacing from $8.9$ to $13.1$ tokens ($+0.22\,\text{dB}$). (\textbf{c}, \textbf{d}) Zero-header $B{=}5{,}632$ observes $100\%$ of Rayleigh tokens via soft LLRs and retains $0.3851$ LPIPS vs.\ $0.6445$ (TONIC) and $0.7163$ (TokenCom+AMC) at $\text{NMSE}{=}0.30$.}
  \label{fig:physim}
\end{figure*}

Because $\lambda_{\max}(M_F)I - M_F \succeq 0$ and $\Sigma(O) \succeq 0$, their symmetric product trace is non-negative ($\operatorname{tr}(\Sigma^{1/2}(\lambda_{\max}(M_F)I - M_F)\Sigma^{1/2}) \ge 0$), so Lemma~\ref{lem:anisotropy} bounds the covariance term via $\operatorname{tr}(M_F\Sigma(O)) \le \lambda_{\max}(M_F)\operatorname{tr}(\Sigma(O)) \le L_J^2\operatorname{tr}(\Sigma(O))$ (where the uniform constant $L_J^2$ applies across observations as $M_F(\mu(O))$ varies, while total perceptual risk additionally includes the bias and remainder terms). Lowering $\operatorname{tr}(\Sigma(O))$ does not guarantee per-eigenvector reduction without Loewner order ($\Sigma_{\text{new}} \preceq \Sigma_{\text{old}}$), but shrinks this Jacobian-free trace bound without decoder gradients, while auxiliary codeword cross-entropy $\mathcal{L}_{\text{CE}}$ counters MSE-induced blur~\cite{yao2025vavae} empirically (Sec.~\ref{sec:ablations}).

\subsection{Soft-Recovery Tier Selection}
\label{sec:convexhull}
We next determine when intermediate channel-coding rates remain Pareto-useful alongside strong anchors and omission under an idealized per-unit additive model (additive unit costs and distortions with independent local tier choices; finite block-rounding and joint-restorer coupling are handled in Sec.~\ref{sec:sender}). Given link state $(\bar{\gamma},\text{NMSE},\nu_{\text{fad}})$, let $d_m \triangleq d_u(a_m)$ denote unit $u$'s expected post-recovery distortion under tier $a_m$ ($d_0 \ge \dots \ge d_M \ge 0$). Under the separable Lagrangian $\min \sum_u [d_u(a_u) + \lambda c_{a_u}]$ ($\lambda > 0$), interior tier $a_m$ ($1 \le m < M$) is uniquely optimal over a non-empty positive price interval iff $(c_m, d_m)$ lies strictly on the lower convex hull:
\begin{equation}
  \min_{0\le j < m}\tfrac{d_u(a_j)-d_u(a_m)}{c_m-c_j} > \max_{m < k \le M}\tfrac{d_u(a_m)-d_u(a_k)}{c_k-c_m},
  \label{eq:convexhull}
\end{equation}
reducing for $M = 2$ to chord dominance $(d_0 - d_1)/c_1 > (d_0 - d_2)/c_2$. Indeed, strict Lagrangian optimality $d_m + \lambda c_m < d_j + \lambda c_j$ ($\forall j < m$) and $d_m + \lambda c_m < d_k + \lambda c_k$ ($\forall k > m$) is equivalent to $\underline{\lambda}_m \triangleq \max_{m<k\le M}(d_m-d_k)/(c_k-c_m) < \lambda < \min_{0\le j<m}(d_j-d_m)/(c_m-c_j) \triangleq \overline{\lambda}_m$ (or $\max(0,\underline{\lambda}_m) < \overline{\lambda}_m$ if monotonicity $d_m \ge d_k$ is relaxed). Because $d_m \ge d_k$ implies $\underline{\lambda}_m \ge 0$, the open interval $(\underline{\lambda}_m,\overline{\lambda}_m) \subset (0,\infty)$ is non-empty (containing midpoint $\lambda = \tfrac{1}{2}(\underline{\lambda}_m+\overline{\lambda}_m) > 0$) iff Eq.~\eqref{eq:convexhull} holds. Crucially, once $M \ge 3$, local adjacent-pair checks ($j=m{-}1, k=m{+}1$) do not suffice: four tiers with costs $(0,1,2,3)$ and distortions $(10,8,7,1)$ satisfy adjacent slopes $(10{-}8)/1 = 2 > (8{-}7)/1 = 1$ at $a_1$, yet the global slope to $a_3$ is $(8{-}1)/2 = 3.5 > 2$, lifting $(1,8)$ above chord $\overline{a_0 a_3}$ so no $\lambda > 0$ ever selects $a_1$. This motivates the isotonic upper-hull pooling heuristic in Sec.~\ref{sec:sender}.

\begin{theorem}[Convex-Hull Separation via Soft Recovery]
\label{thm:unfolding}
Consider $\{a_0 = \textup{\textsc{Drop}}, a_1 = \textup{\textsc{Weak}}, a_2 = \textup{\textsc{Strong}}\}$ with $0 = c_0 < c_1 < c_2$. For a unit whose block-level pass/fail indicator has failure probability $\epsilon_m \triangleq p_{\textup{fail}}(a_m) \in [0,1)$ ($m \in \{1,2\}$), assume conditional expectations of the same unit distortion satisfy a shared verified distortion $d_u^{\textup{succ}} < d_u(a_0)$ and failed risk $d_u^{\textup{fail}}(a_m) \in [d_u^{\textup{succ}}, d_u(a_0)]$. With soft recovery efficiency $\eta_{u,m} \in [0,1]$,
\begin{equation}
  \eta_{u,m} \triangleq \frac{d_u(a_0)-d_u^{\textup{fail}}(a_m)}{d_u(a_0)-d_u^{\textup{succ}}},
  \label{eq:eta}
\end{equation}
$(c_1, d_u(a_1))$ lies strictly below chord $\overline{a_0 a_2}$ if and only if
\begin{equation}
  1-\epsilon_1(1-\eta_{u,1}) > \frac{c_1}{c_2}\big(1-\epsilon_2(1-\eta_{u,2})\big).
  \label{eq:unfolding}
\end{equation}
With $r \triangleq c_1/c_2$, hard erasure ($\eta_{u,1} = 0$) loses strict chord dominance whenever $\epsilon_1 \ge 1 - r + r\epsilon_2(1-\eta_{u,2})$ (reducing to $\epsilon_1 \ge 1 - r$ in the reliable-\textup{\textsc{Strong}} limit $\epsilon_2 \to 0$). Conversely, when $\epsilon_2 \to 0$ and $\epsilon_1 > 0$, soft recovery with $\eta_{u,1} > 1-(1-r)/\epsilon_1$ ($< 1$ since $r < 1$) strictly places \textup{\textsc{Weak}} on the three-tier lower hull.
\end{theorem}

\begin{proof}
Under the unit pass/fail decomposition, $d_u(a_m) = (1-\epsilon_m)d_u^{\text{succ}} + \epsilon_m d_u^{\text{fail}}(a_m)$. Subtracting from $d_u(a_0)$ and substituting Eq.~\eqref{eq:eta} gives $d_u(a_0) - d_u(a_m) = (d_u(a_0) - d_u^{\text{succ}})\,[1 - \epsilon_m(1 - \eta_{u,m})]$. Substituting this expression for $m \in \{1,2\}$ into the $M=2$ hull criterion $(d_u(a_0)-d_u(a_1))/c_1 > (d_u(a_0)-d_u(a_2))/c_2$ and canceling $d_u(a_0)-d_u^{\text{succ}} > 0$ yields Eq.~\eqref{eq:unfolding}. Setting $\eta_{u,1}=0$ in Eq.~\eqref{eq:unfolding} gives $1-\epsilon_1 > r(1-\epsilon_2(1-\eta_{u,2}))$, which fails once $\epsilon_1 \ge 1 - r + r\epsilon_2(1-\eta_{u,2})$ (at $\epsilon_1 = 1-r$ with $\eta_{u,2}<1$, any $\epsilon_2 > 0$ still keeps \textsc{Weak} strictly below the chord, reaching collinearity as $\epsilon_2 \to 0$). In the $\epsilon_2 \to 0$ limit, solving $1-\epsilon_1(1-\eta_{u,1}) > r$ for $\eta_{u,1}$ when $\epsilon_1 > 0$ gives $\eta_{u,1} > 1 - (1-r)/\epsilon_1 < 1$.
\end{proof}

Even when intermediate tiers occupy the lower hull, signaling per-unit schedules incurs control-header overhead that can outweigh dynamic gains near full token coverage.

\begin{corollary}[Tier Monotonicity and Header Overhead]
\label{cor:granularity}
With $M+1$ tiers, $B_{\textup{tot}} = B_{\textup{pay}}(M+1)+B_{\textup{anchor}}+\Omega(M+1)$ splits budget across payload, anchors, and non-decreasing incremental header/framing overhead $\Omega(M+1)$ normalized relative to the zero-header baseline ($\Omega(1) = 0$). For nested tier sets $\mathcal{A}_M \subseteq \mathcal{A}_{M+1}$ and fixed $\Omega$, optimal integer distortion satisfies $D_{M+1}^\star(B) \le D_M^\star(B)$, and the convexified LP relaxation $\tilde{D}_{M+1}^\star(B) \le \tilde{D}_M^\star(B)$ achieves equality $\tilde{D}_{M+1}^\star(B) = \tilde{D}_M^\star(B)$ whenever all added tiers fail to strictly lower $\min_{a \in \mathcal{A}_M}(d_u(a)+\lambda c_a)$ for any $\lambda \ge 0$ (integer schedules can still differ by knapsack slack, and empirical reversals stem from $\Omega$ jumps or finite framing/optimization). On the convexified envelope with local shadow price $\lambda^\star > 0$, omitting headers strictly lowers envelope distortion whenever $\Delta\tilde{D}_{\textup{alloc}} < \lambda^\star\,\Delta\Omega$.
\end{corollary}

\begin{proof}
Let $B_{\text{net}} \triangleq B_{\text{tot}} - B_{\text{anchor}}$ denote the post-anchor budget. At fixed framing cost $\Omega$, nested feasibility $\mathcal{P}_M \subseteq \mathcal{P}_{M+1}$ immediately yields $D_{M+1}^\star(B_{\text{pay}}) \le D_M^\star(B_{\text{pay}})$. Define the convexified relaxation $\tilde{D}_T^\star(B)$ as the finite linear program minimizing $\sum_{u=1}^U \sum_{a \in \mathcal{A}_T} p_{u,a} d_u(a)$ over per-unit probability mixtures $p_u \in \Delta(\mathcal{A}_T)$ subject to $\sum_{u=1}^U \sum_{a \in \mathcal{A}_T} p_{u,a} c_a \le B$. By LP strong duality, $\tilde{D}_T^\star(B) = \sup_{\lambda \ge 0}\big[\sum_{u=1}^U \min_{a \in \mathcal{A}_T}(d_u(a) + \lambda c_a) - \lambda B\big]$ is non-increasing and convex in $B$. Whenever all tiers in $\mathcal{A}_{M+1}\setminus\mathcal{A}_M$ lie on or above the lower convex hull of $\mathcal{A}_M$, $\min_{a \in \mathcal{A}_{M+1}}(d_u(a)+\lambda c_a) = \min_{a \in \mathcal{A}_M}(d_u(a)+\lambda c_a)$ for every $\lambda \ge 0$ (extending to $\lambda = 0$ by continuity on finite $\mathcal{A}_T$), yielding $\tilde{D}_{M+1}^\star(B_{\text{pay}}) = \tilde{D}_M^\star(B_{\text{pay}})$ (whereas exact integer optima $D_{M+1}^\star, D_M^\star$ may still differ by finite knapsack rounding). Next, define the overhead jump $\Delta\Omega \triangleq \Omega(M+1) - \Omega(1) = \Omega(M+1) > 0$ (giving $B_{\text{pay}}(1) = B_{\text{net}}$) and the gross fixed-overhead allocation gain $\Delta\tilde{D}_{\text{alloc}} \triangleq \tilde{D}_1^\star(B_{\text{net}}) - \tilde{D}_{M+1}^\star(B_{\text{net}}) \ge 0$. Since $\tilde{D}_{M+1}^\star(B)$ is non-increasing and convex in $B$, any supporting shadow price $-\lambda^\star \in \partial \tilde{D}_{M+1}^\star(B_{\text{net}})$ ($\lambda^\star > 0$ on non-saturated budgets) satisfies $\tilde{D}_{M+1}^\star(B_{\text{net}}-\Delta\Omega) - \tilde{D}_{M+1}^\star(B_{\text{net}}) \ge \lambda^\star\,\Delta\Omega$. Telescoping around $B_{\text{net}}$ gives
\begin{equation}
  \begin{aligned}
    &\tilde{D}_{M+1}^\star(B_{\text{net}}\!-\!\Delta\Omega) - \tilde{D}_1^\star(B_{\text{net}}) \\
    &\quad = \big[\tilde{D}_{M+1}^\star(B_{\text{net}}\!-\!\Delta\Omega) - \tilde{D}_{M+1}^\star(B_{\text{net}})\big] \\
    &\qquad - \Delta\tilde{D}_{\text{alloc}}
    \;\ge\; \lambda^\star\,\Delta\Omega - \Delta\tilde{D}_{\text{alloc}},
  \end{aligned}
  \label{eq:ext_overhead_trade}
\end{equation}
so omitting the header strictly lowers envelope distortion whenever $\Delta\tilde{D}_{\text{alloc}} < \lambda^\star\,\Delta\Omega$ (and directly implies integer preference whenever the headerless baseline has zero integrality gap).
\end{proof}

These principles directly shape AdapToC: soft recovery preserves intermediate tiers on the three-tier hull (Sec.~\ref{sec:receiver}), while deterministic zero-header switching returns all $384$ header symbols to payload near full coverage (Sec.~\ref{sec:sender}; Fig.~\ref{fig:physim}).

\section{AdapToC}
\label{sec:method}

We now translate the rate--distortion conditions of Sec.~\ref{sec:theory} into a concrete single-pass transceiver. Given image $I$, frozen $(E,\mathcal{C},D)$, link state $(\bar{\gamma},\nu_{\text{fad}},\text{NMSE})$, and budget $B_{\text{tot}}$, AdapToC compacts and bit-aligns tokens, schedules tiers, and restores $\hat{Z} \in \mathbb{R}^{L\times d_z}$ in one pass for either pixel synthesis $D(\hat{Z})$ or latent task heads (Fig.~\ref{fig:overview}a).

\subsection{Codebook Alignment and Spatial Ordering}
\label{sec:geometry}

From VQGAN-$f_{16}$ ($16{,}384$ entries), we retain active codes $\mathcal{C}_{\text{act}}$ ($K = 2^b = 1{,}024$, $b = 10$) profiled strictly on training data ($126{,}982$ images) and shorten raw indices by $28.6\%$ ($14 \to 10$ bits). Freezing $\mathcal{C}_{\text{act}}$, empirical codeword frequencies $w_{\pi(g)}$, and prefix tables prior to evaluation achieves a $100.0\%$ hit rate ($0\%$ remapping) on $2{,}865{,}024$ ImageNet tokens ($4{,}974$ leakage-free samples from $5{,}000$ after removing $26$ train overlaps) and $589{,}824$ NuPlan tokens; all same-tokenizer baselines share this representation. With bitwise XOR $\oplus$, Hamming distance $d_H$, and $r$-th basis vector $e_r \in \{0,1\}^b$, bijection $\pi:\{0,1\}^b \to \mathcal{C}_{\text{act}}$ iteratively decreases the frequency-weighted $1$-bit energy $\mathcal{E}(\pi) = \sum_{g,r}w_{\pi(g)}\|\pi(g)-\pi(g\oplus e_r)\|_2^2$ via bisection and conflict-free parallel $2$-Opt sweeps (Proposition~\ref{prop:hypercube}).

\begin{proposition}[Conflict-Free Hypercube Matching]
\label{prop:hypercube}
For $b \ge 2$, coordinate $b^\star \in \{1,\dots,b\}$, and parity $p \in \{0,1\}$, define the half-hypercube
\begin{equation}
  V_{b^\star, p} = \Big\{ g \in \{0,1\}^b \;\Big|\; \bigoplus_{r \neq b^\star} g_r = p \Big\},
  \label{eq:hypercube}
\end{equation}
with $|V_{b^\star,p}| = 2^{b-1}$. In matching $\mathcal{M}_{b^\star, p} \triangleq \{\{g, g\oplus e_{b^\star}\} \mid g \in V_{b^\star, p}, g_{b^\star} = 0\}$, each pair forms a $d_H = 1$ edge, whereas distinct pairs satisfy $d_H \ge 2$ with zero $1$-bit cross-edges, so parallel swaps decouple additively.
\end{proposition}

\begin{proof}
First, because $b \ge 2$ ensures at least one index $r \neq b^\star$ in the XOR sum, $|V_{b^\star,p}| = 2^{b-1}$ for each $p \in \{0,1\}$, and flipping coordinate $b^\star$ leaves $\bigoplus_{r\neq b^\star} g_r$ in Eq.~\eqref{eq:hypercube} invariant, so $g \in V_{b^\star,p} \iff g \oplus e_{b^\star} \in V_{b^\star,p}$ and every pair $\{g,g'\} \in \mathcal{M}_{b^\star,p}$ ($g' = g \oplus e_{b^\star}$) is a valid internal edge of Hamming distance $d_H(g,g') = 1$. Second, take any two distinct pairs $\{g,g'\}$ and $\{\tilde{g},\tilde{g}'\}$ in $\mathcal{M}_{b^\star,p}$; if some $g^{(1)} \in \{g,g'\}$ and $g^{(2)} \in \{\tilde{g},\tilde{g}'\}$ had $d_H(g^{(1)},g^{(2)}) = 1$, they would disagree on a single coordinate $r \neq b^\star$ ($g^{(2)} = g^{(1)} \oplus e_r$), flipping the summation parity to $\bigoplus_{j\neq b^\star} g^{(2)}_j = 1 - p$ and contradicting $g^{(2)} \in V_{b^\star,p}$. Thus distinct pairs are separated by $d_H \ge 2$ with zero $1$-bit cross-edges. Finally, along internal edge $\{g,g'\}$, swapping $\pi(g)$ and $\pi(g')$ leaves $(w_{\pi(g)}+w_{\pi(g')})\|\pi(g)-\pi(g')\|_2^2$ symmetric and invariant, while every cross-edge $\{g^{(1)}, g^{(1)}\oplus e_r\}$ ($r \neq b^\star$) crosses into the fixed complement $V_{b^\star,1-p}$. Swapping $\{g,g'\}$ therefore affects only edges touching $V_{b^\star,1-p}$ and leaves the swap delta of every other pair in $\mathcal{M}_{b^\star,p}$ unchanged, so $\Delta\mathcal{E}(\pi)$ sums the $2^{b-2}$ pair deltas exactly.
\end{proof}

Sweeping $(b^\star,p)$ cuts uniform-edge $d_H = 1$ MSE across $1{,}024$ VQGAN codes by $61.1\%$ ($1.549 \to 0.603$; Fig.~\ref{fig:physim}a) while preserving marginal codebook variance. Canonical prefix coding packs $11$ tokens per $128$-bit block ($112$ payload bits $+$ CRC-$16$), or $12$ when $\sum \ell_{\text{pfx}}(g_i) \le 112$ (indexed by $24$ Block-$0$ flags covering the first $24$ \textsc{Weak} blocks), and parses decoded bits via a prefix lookup table (LUT) to reconstruct $10$-bit Gray pseudo-LLRs $(2\hat{g}_{i,r}-1)\max(\min_j |\boldsymbol{\ell}_{i,j}^{\text{pfx}}|, 10^{-4})$ ($0$ if truncated) before $(1+3\,\text{NMSE})^{-1}$ damping. Around the $36$ anchors $G_0$, each tier permutes its $6\times 6$ grid via a rotating 2D Halton key (Fig.~\ref{fig:physim}b) to widen intra-block spacing ($8.9 \to 13.1$ mean tokens; $2.12 \to 3.73$ min) at zero bit cost, leaving more surviving $3\times 3$ neighbors around burst-erased tokens to assist local spatial inpainting.

\subsection{Budget-Regime Tier Allocation}
\label{sec:sender}

Sender $S_\theta$ ($8.16\,\text{M}$ params; Fig.~\ref{fig:overview}b) embeds $L = 576$ tokens via three ConvNeXt-SwiGLU blocks, $4\times 4$ pooling ($U = 36$), and two cross-scale layers into $H^{(0)} \in \mathbb{R}^{U\times d}$. With nominal per-token costs $c_1 \equiv c_W$ (\textsc{Weak}) and $c_2 \equiv c_S$ (\textsc{Strong}), header $B_{\text{hdr}} \in \{0,384\}$, and relaxed non-anchor budget $B_{\text{free}} = B_{\text{tot}}-B_{\text{hdr}}-36\,c_2$, six unit-gated AdaLN blocks ($l \in \{1,\dots,6\}$, Hadamard product $\odot$; LN, MHA, and SwiGLU preserve $\mathbb{R}^{U\times d}$ while AdaLN vectors in $\mathbb{R}^d$ broadcast across units) condition on $\mathbf{s}_{\text{link}} = [\bar{\gamma}/10,\allowbreak \nu_{\text{fad}},\allowbreak \text{NMSE},\allowbreak B_{\text{free}}/(L c_2),\allowbreak B_{\text{free}}/(L c_1),\allowbreak c_1/c_2]^\top$ via zero-initialized multi-layer perceptrons $(\boldsymbol{\gamma}_{j,l},\boldsymbol{\beta}_{j,l},\boldsymbol{\alpha}_{j,l})_{j=1}^2 = \operatorname{MLP}_l(\mathbf{s}_{\text{link}})$ for $j \in \{1,2\}$:
\begin{equation}
  \begin{aligned}
    &\phi_{l,j}(X) = (1+\boldsymbol{\gamma}_{j,l})\odot\operatorname{LN}(X)+\boldsymbol{\beta}_{j,l}, \\[-1pt]
    &\tilde{H}^{(l)} = H^{(l-1)} + (1+\boldsymbol{\alpha}_{1,l})\odot\operatorname{MHA}(\phi_{l,1}(H^{(l-1)})), \\[-1pt]
    &H^{(l)} = \tilde{H}^{(l)} + (1+\boldsymbol{\alpha}_{2,l})\odot\operatorname{SwiGLU}(\phi_{l,2}(\tilde{H}^{(l)})),
  \end{aligned}
  \label{eq:sender_adaln}
\end{equation}
where zero-initializing $\boldsymbol{\alpha}_{j,l} = 0$ in Eq.~\eqref{eq:sender_adaln} preserves residual identity skips $(1+\boldsymbol{\alpha}_{j,l}) = 1$ at initialization and passes immediate gradients from $B_{\text{free}}$.

\begin{figure*}[t]
  \centering
  \setlength{\abovecaptionskip}{2pt}
  \setlength{\belowcaptionskip}{1pt}
  \includegraphics[width=0.95\linewidth]{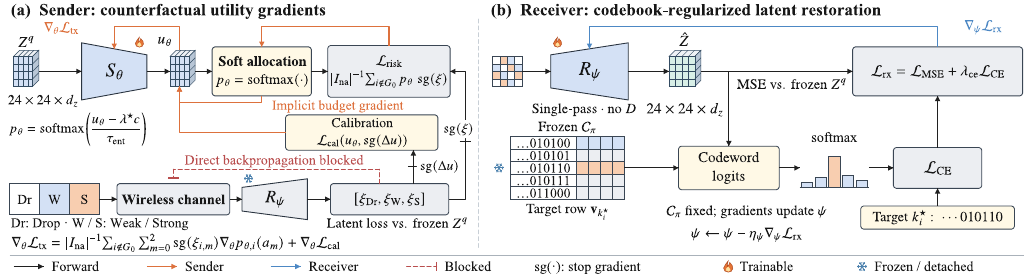}
  \caption{\textbf{Latent co-training aligns allocation and restoration without decoder gradients.} (\textbf{a}) Coupled rollouts under $\operatorname{sg}(\xi)$ train $S_\theta$ via $\mathcal{L}_{\text{risk}}$ and $\mathcal{L}_{\text{cal}}$. (\textbf{b}) Targets $Z^q$ and codeword cross-entropy $\mathcal{L}_{\text{CE}}$ regularize $R_\psi$ in latent space.}
  \label{fig:training}
\end{figure*}

Each $4\times 4$ unit $u$ partitions its phases into anchor $G_{u,0} = \{t_0\}$ (\textsc{Strong}, forming global anchor set $G_0 = \bigcup_u G_{u,0}$, $|G_0|=36$) and non-anchor phases $G_{u,1} = \{t_1,\dots,t_3\}$, $G_{u,2} = \{t_4,\dots,t_7\}$, and $G_{u,3} = \{t_8,\dots,t_{15}\}$. A Softplus head predicts non-negative token gains $(u_{\theta,i}(a_1), u_{\theta,i}(a_2))$ ($u_{\theta,i}(a_0) = 0$) pooled on $G_{u,1},\dots,G_{u,3}$. A global $2$-bit split $\kappa \in \{0,\dots,3\}$ assigns levels $1,\dots,\kappa$ to \textsc{Strong} and $\kappa+1,\dots,3$ to \textsc{Weak}; isotonic upper-hull pooling across levels $1,\dots,3$ serves as a convex-hull relaxation ranking heuristic inspired by Eq.~\eqref{eq:convexhull} (while noting that exact integer prefix selection can still include pooled intermediate depths under tight budgets), solved via $32$-step Newton bisection in training or greedy integer depths $d_u \in \{0,\dots,3\}$ ($72$ bits, distinct from distortion $d_u(a_m)$) across $\kappa$ at inference. Quadrature phase-shift keying (QPSK) uses $R = 2/3$ LDPC for \textsc{Weak} ($96$ sym/blk, nominal $c_1 \in [8.0,8.7]$ sym/tok) and $N_{\text{sym}}^S \in \{192,96\}$ for \textsc{Strong} ($R = 1/3$, $c_2 \approx 17.5$ in fading; $R = 2/3$ on clean AWGN, collapsing $c_1=c_2$ to \textsc{Drop}/\textsc{Weak}). For $n_S \ge 36$ \textsc{Strong} and $n_W$ \textsc{Weak} tokens with $N_{\text{blk}}^W(n_W) = K_{12}+\lceil\max(0,n_W-12K_{12})/11\rceil$ ($K_{12} \le \min(24,\lfloor n_W/12\rfloor)$ over the first $24$ \textsc{Weak} blocks), every transmitted QPSK frame satisfies the symbol ledger (excluding pilots; optional $16$-QAM \textsc{Weak} uses $64$ sym/blk, and headered mode requires $B_{\text{tot}} \ge 1{,}152$ in fading or $768$ on clean AWGN to carry $36$ anchors plus Block-$0$)
\begin{equation}
  B_{\text{used}} = B_{\text{hdr}} + N_{\text{sym}}^S \Big\lceil \frac{n_S}{11} \Big\rceil + 96\,N_{\text{blk}}^W(n_W) \le B_{\text{tot}}.
  \label{eq:ledger}
\end{equation}
Both sender and receiver first determine $N_{\text{sym}}^S$ from shared link statistics ($N_{\text{sym}}^S = 96$ iff $\nu_{\text{fad}} \le 0.05$ and $\hat{\gamma}_{\text{eff,dB}} \ge 6 + 25\nu_{\text{fad}}$, else $192$), fixing $B_{\text{full}} = 4N_{\text{sym}}^S + 50\times 96$ ($5{,}568$ fading; $5{,}184$ clean) in Eq.~\eqref{eq:ledger}, and set the deterministic re-entry threshold $B_{\text{up}} = B_{\text{full}} + 480$ ($6{,}048$ fading), an empirical protocol margin above $B_{\text{full}} + B_{\text{hdr}}$ ($5{,}952$) near the surplus transition (Sec.~\ref{sec:ablations}). Guided by Corollary~\ref{cor:granularity}, below $B_{\text{full}}$ (deficit) and at or above $B_{\text{up}}$ (surplus), Block-$0$ ($B_{\text{hdr}} = 384$, $384\text{h}$) signals the adaptive tier map, whereas inside $[B_{\text{full}},B_{\text{up}})$ both ends deterministically omit headers ($0\text{h}$: $36$ \textsc{Strong} anchors, $540$ fixed $11$-tok \textsc{Weak}). Block-$0$ packs $128$ bits ($101$ control bits: $72$ unit depth, $2$ split $\kappa$, $3$ PHY mode/MCS, $24$ $12$-token flags; plus $11$ fixed zero-padding bits and CRC-$16$), encoded by $R = 1/3$ LDPC into $384$ coded bits ($192$ QPSK symbols) and repeated twice at $\text{AL} = 2$ with soft combining for $B_{\text{hdr}} = 384$ symbols, reserving Bit-$74$ for an optional fourth $16$-QAM $R = 1/2$ \textsc{Weak} rung ($64$ sym/$12$-tok block; Sec.~\ref{sec:ablations}).

\subsection{Pilot-Calibrated Soft Recovery}
\label{sec:receiver}

At the receiver, $R_\psi$ fuses CRC-verified anchors, CRC-failed soft posteriors, and omitted positions in one pass (Fig.~\ref{fig:overview}c). Proposition~\ref{prop:llr} isolates why noisy pilots distort uncalibrated demappers and gives the exact effective noise variance used before posterior readout.

\begin{proposition}[Pilot-Error LLR Inflation]
\label{prop:llr}
Under $y_s = h_s x_s + n_s$ with jointly circularly symmetric complex Gaussian fading and pilot noise, independent data $x_s$ and thermal noise $n_s \sim \mathcal{CN}(0,\sigma_n^2)$ ($\sigma_n^2 > 0$), orthogonal LMMSE estimate $\hat{h}_s = h_s - e_s$ ($e_s \sim \mathcal{CN}(0,\sigma_{e,s}^2)$, $\mathbb{E}[\hat{h}_s e_s^*] = 0$), and unit-modulus Gray-mapped QPSK ($|x_s|^2 = 1$) encoding independent equiprobable bits, the effective noise $w_s \triangleq y_s - \hat{h}_s x_s = e_s x_s + n_s$ conditioned on $(\hat{h}_s, x_s)$ has variance $\sigma_{\textup{eff},s}^2 = \sigma_n^2 + \sigma_{e,s}^2$; scaling per-symbol marginal demapper LLRs by $\sigma_n^2$ alone satisfies $\ell_s^{\textup{naive}} = \tau_s^\star \ell_s^{\textup{exact}}$ with exact inflation factor $\tau_s^\star = 1 + \sigma_{e,s}^2/\sigma_n^2$.
\end{proposition}

\begin{proof}
Under jointly circularly symmetric complex Gaussian fading and pilot noise, LMMSE orthogonality $\mathbb{E}[\hat{h}_s e_s^*] = 0$ implies independence of $e_s$ and $\hat{h}_s$; since $x_s$ and $n_s$ are independent of $(\hat{h}_s, e_s)$ and $|x_s|^2 = 1$, conditioning on $(\hat{h}_s, x_s)$ yields $w_s \sim \mathcal{CN}(0,\sigma_{\text{eff},s}^2)$ with $\sigma_{\text{eff},s}^2 = \sigma_n^2 + \sigma_{e,s}^2$ (splitting into two independent real components of variance $\sigma_{\text{eff},s}^2/2$). Note that this exactness holds for per-symbol marginal likelihoods (when a single pilot error $e_s$ is shared across a symbol vector $x$, the joint conditional noise covariance is $\sigma_n^2 I + \sigma_{e,s}^2 x x^{\mathrm{H}}$). For Gray-mapped QPSK $x_s = (s_I + \mathrm{j} s_Q)/\sqrt{2}$ encoding independent equiprobable bits $(\beta_I, \beta_Q) \in \{0,1\}^2$, when $\hat{h}_s = 0$ we have $\ell_s^{\text{exact}} = \ell_s^{\text{naive}} = 0$; when $\hat{h}_s \neq 0$, phase-rotating by $\hat{h}_s^*/|\hat{h}_s|$ factorizes $P(y_s \mid \hat{h}_s, \beta_I, \beta_Q)$ into independent in-phase and quadrature Gaussians. Marginalizing $\ell_{s,I}^{\text{exact}} = \log\big[\sum_{\beta_Q} P(y_s\mid\hat{h}_s,\beta_I{=}1,\beta_Q)/\sum_{\beta_Q} P(y_s\mid\hat{h}_s,\beta_I{=}0,\beta_Q)\big]$ cancels the perpendicular bit factor identically with zero Max-Log approximation, leaving $\ell_s^{\text{exact}} = (d_0^2 - d_1^2)/\sigma_{\text{eff},s}^2$ (where $d_0^2, d_1^2$ are the active-axis constellation distances) versus $\ell_s^{\text{naive}} = (d_0^2 - d_1^2)/\sigma_n^2$, hence $\ell_s^{\text{naive}} = (1 + \sigma_{e,s}^2/\sigma_n^2)\,\ell_s^{\text{exact}} = \tau_s^\star \ell_s^{\text{exact}}$ (including on decision axes where $\ell_s^{\text{exact}} = 0$). Moreover, across heterogeneous subband CSI errors, Cauchy--Schwarz shows that unbiased linear diversity combining achieves minimum conditional symbol MSE using effective-precision weights $\alpha_s^\star \propto |\hat{h}_s|^2/\sigma_{\text{eff},s}^2 \propto \hat{\gamma}_s$.
\end{proof}

Physical demapping scales directly by $\sigma_{\text{eff},s}^2$, and soft readout damps finite-length LDPC LLRs via $\tilde{\boldsymbol{\ell}}_i = (1+3\,\text{NMSE})^{-1}\boldsymbol{\ell}_i$ (Fig.~\ref{fig:physim}d), using bit-LLR convention $\ell_{i,r} = \log[P(g_{i,r}=1)/P(g_{i,r}=0)]$. With codeword matrix $G \in \{0,1\}^{K\times b}$ ($K = 2^b$, $b \in \{10,12\}$, row $g$ equal to $g^\top$), factorizing $\tilde{\boldsymbol{\ell}}_i$ over bits under a uniform prior on $\mathcal{C}_{\text{act}}$ yields posterior $\mathbf{q}_i = [q_i(g)]_{g\in\{0,1\}^b} \in \mathbb{R}^K$ and expectation $\bar{z}_i \in \mathbb{R}^{d_z}$ (distinct from Lemma~\ref{lem:anisotropy}'s exact joint $\mu(O)$; $\tilde{\boldsymbol{\ell}}_i = 0$ reverts to the centroid):
\begin{equation}
  \mathbf{q}_i = \operatorname{softmax}\!\big(G\tilde{\boldsymbol{\ell}}_i\big), \qquad \bar{z}_i = \sum_{g\in\{0,1\}^b} q_i(g)\,\pi(g).
  \label{eq:readout}
\end{equation}
When $g_i^{\text{sent}} = 0$ (which implies $g_i^{\text{crc}} = 0$), $R_\psi$ inserts a learnable omission embedding $e_{\text{omit}} \in \mathbb{R}^{d_z}$ while zeroing $\bar{z}_i$ from Eq.~\eqref{eq:readout} in Eq.~\eqref{eq:receiver}. Parameter-free \emph{AdapToC-PnP} outputs $\hat{z}_i^{\text{PnP}} = g_i^{\text{crc}}\pi(\hat{g}_i) + (1-g_i^{\text{crc}})g_i^{\text{sent}}\bar{z}_i$ (equal to $z_i^q$ when $g_i^{\text{crc}}=1$ absent undetected CRC errors) and transfers directly to 1D TiTok~\cite{yu2024titok} and multi-scale VAR~\cite{tian2024var}.

For 2D spatial grids, restorer $R_\psi$ ($3.52\,\text{M}$ params) projects $[\hat{z}_i^{\text{PnP}} + (1-g_i^{\text{sent}})e_{\text{omit}}, \mathbf{r}_i]$ plus spatial embeddings into $X^{(0)} \in \mathbb{R}^{L\times d_{\text{rx}}}$. Feature vector $\mathbf{r}_i \in \mathbb{R}^{21}$ packs $14$-slot $|\boldsymbol{\ell}_i|/12$ ($4$ trailing zero-bit pad slots at $|-24|/12 = 2.0$ for $10$-bit VQGAN/$14$-bit ablation sharing when $g_i^{\text{sent}}=1$ and $0$ when omitted; $12$-bit TiTok/VAR use dedicated restorers, Sec.~\ref{sec:limitations}), flags $(g_i^{\text{crc}}, g_i^{\text{sent}})$, mean/min $|\boldsymbol{\ell}_i|/12$ over the $14$ slots, $\bar{\gamma}/10$, $\text{NMSE}$, and $g_i^{\text{sent}}\hat{\gamma}_{s(i),\text{dB}}/10$ (zeroed on omitted tokens and floored when $\hat{h}_s=0$). Three $3\times 3$ ConvNeXt-SwiGLU blocks~\cite{liu2022convnext} inpaint omissions before four attention layers (with $\mathbf{q}_{h,i},\mathbf{k}_{h,j} \in \mathbb{R}^{d_h}$, zero-initialized $\mathbf{w}_h \in \mathbb{R}^{21}$, and $A_{ij}^{(h)} = \mathbf{q}_{h,i}^\top\mathbf{k}_{h,j}/\sqrt{d_h} + \mathbf{w}_h^\top\mathbf{r}_j$) predict $\Delta_{\psi}(X^{(0)}) \in \mathbb{R}^{L\times d_z}$ under exact CRC bypass:
\begin{equation}
  \hat{z}_i = g_i^{\text{crc}}\,\pi(\hat{g}_i) + \big(1-g_i^{\text{crc}}\big)\big(g_i^{\text{sent}}\bar{z}_i + \Delta_{\psi}(X^{(0)})_i\big).
  \label{eq:receiver}
\end{equation}

\subsection{Decoder-Free Latent Co-Training}
\label{sec:cotraining}

Joint co-training couples $S_\theta$ and $R_\psi$ without differentiating through frozen pixel decoders in training or inference (Fig.~\ref{fig:training}). With target indices $\mathbf{k}^\star = (k_1^\star,\dots,k_L^\star) \in \{0,\dots,K-1\}^L$ ($z_i^q = v_{k_i^\star}$), codeword cross-entropy $\ell_{\text{CE}}(\hat{z}_i,k_i^\star) \triangleq -\log\operatorname{softmax}_k(-\|\hat{z}_i-v_k\|_2^2/\tau_{\text{ce}})_{k_i^\star}$ ($\tau_{\text{ce}} = 25.0$), and $\mathcal{L}_{\text{CE}}(\hat{Z},\mathbf{k}^\star) \triangleq L^{-1}\sum_{i=1}^L\ell_{\text{CE}}(\hat{z}_i,k_i^\star)$ (empirically countering MSE-induced blur without $J_F$; Sec.~\ref{sec:ablations}), $R_\psi$ minimizes Frobenius MSE plus $\mathcal{L}_{\text{CE}}$ ($\lambda_{\text{ce}} = 0.02$; Fig.~\ref{fig:training}b):
\begin{equation}
  \mathcal{L}_{\text{rx}}(\psi) = (Ld_z)^{-1}\|\hat{Z}-Z^q\|_F^2 + \lambda_{\text{ce}}\,\mathcal{L}_{\text{CE}}(\hat{Z}, \mathbf{k}^\star).
  \label{eq:rxloss}
\end{equation}
Because mixed schedules couple units via $d_i(a_i,\mathbf{a}_{-i})$, paired uniform non-anchor rollouts $a_m \in \{a_0,a_1,a_2\}$ ($O(3)$ passes; Fig.~\ref{fig:training}a) yield restored $\hat{z}_{i,m}$ as a low-variance utility proxy validated under full mixed perturbations (Sec.~\ref{sec:ablations}). With rollout risk $\xi_{i,m} \triangleq d_z^{-1}\|\hat{z}_{i,m}-z_i^q\|_2^2 + \lambda_{\text{ce}}\ell_{\text{CE}}(\hat{z}_{i,m}, k_i^\star)$, calibrated target $\Delta u_{i,m} = \max(0,\xi_{i,0}-\xi_{i,m})$, stop-gradient $\operatorname{sg}(\cdot)$, Smooth-$L_1$ loss $\mathcal{L}_{1,\text{sm}}$ ($\beta_{\text{cal}} = 1.0$), and entropy-regularized Gibbs policy $p_{\theta,i}(a_m) = \operatorname{softmax}_m((u_{\theta,i}(a_m)-\lambda^\star c_m)/\tau_{\text{ent}})$ ($\tau_{\text{ent}} = 0.15$, backpropagated via the implicit dual gradient of $\lambda^\star(\theta)$ when active), $S_\theta$ minimizes
\begin{equation}
  \begin{aligned}
    &\mathcal{L}_{\text{tx}}(\theta) = \frac{1}{L-|G_0|}\sum_{i\notin G_0} \Big[\sum_{m=0}^2 p_{\theta,i}(a_m)\operatorname{sg}(\xi_{i,m}) \\[-2pt]
    &\qquad + \frac{\beta_{\text{cal}}}{2}\sum_{m=1}^2 \mathcal{L}_{1,\text{sm}}\!\big(u_{\theta,i}(a_m), \operatorname{sg}(\Delta u_{i,m})\big)\Big],
  \end{aligned}
  \label{eq:txloss}
\end{equation}
where $L - |G_0| = 540$ counts the non-anchor tokens.

\paragraph{Dual optimality and cardinal anchoring.}
Equation~\eqref{eq:txloss} pairs unique dual budget enforcement with cardinal scale calibration. For predicted gains $u_{\theta,i}(a_m) \ge 0$, minimizing the strictly convex entropy-regularized Lagrangian $\sum_{i \notin G_0}\sum_{m=0}^2 [p_i(a_m)(\lambda c_m - u_{\theta,i}(a_m)) + \tau_{\text{ent}}\,p_i(a_m)\log p_i(a_m)]$ over simplex policies $\{p_i \in \Delta^2\}$ yields the closed-form Gibbs policy $p_{\theta,i}(a_m;\lambda)$. Differentiating the expected non-anchor usage $C(\lambda) \triangleq \sum_{i\notin G_0}\sum_{m=0}^2 p_{\theta,i}(a_m;\lambda)c_m$ gives
\begin{equation}
  \frac{\partial C(\lambda)}{\partial \lambda}
  = -\frac{1}{\tau_{\text{ent}}} \sum_{i \notin G_0} \operatorname{Var}_{p_{\theta,i}(\cdot;\,\lambda)}(c) < 0
  \label{eq:ext_dual_mono}
\end{equation}
for all finite $\lambda$ (since $\tau_{\text{ent}} > 0$ and $0 = c_0 < c_1 < c_2$ prevent degeneracy), with $\lim_{\lambda \to \infty} C(\lambda) = 0$. For any $B_{\text{free}} > 0$ ($B_{\text{free}}=0$ corresponds to all-\textsc{Drop} as $\lambda\to\infty$; $B_{\text{free}}<0$ is infeasible), a unique active multiplier $\lambda^\star > 0$ (when $C(0) > B_{\text{free}}$, else $\lambda^\star = 0$) exists and is bracketed on $[0, \lambda_{\text{hi}}]$ with $\lambda_{\text{hi}} = \big(\max_{i,m} u_{\theta,i}(a_m) + \tau_{\text{ent}}\log(|\mathcal{A}_+|(L-|G_0|)c_{\max}/B_{\text{free}})\big)/c_{\min}$ (accounting for all $|\mathcal{A}_+|$ positive-cost tiers so $C(\lambda_{\text{hi}}) \le B_{\text{free}}$), converging to bracket width $\le 2^{-32}\lambda_{\text{hi}}$ under $32$-step safeguarded Newton bisection. Furthermore, since $\sum_{m=0}^2 p_{\theta,i}(a_m) = 1$, the policy-risk sum rewrites identically as $\sum_{m=0}^2 p_{\theta,i}(a_m)\xi_{i,m} = \xi_{i,0} - \sum_{m=1}^2 p_{\theta,i}(a_m)(\xi_{i,0}-\xi_{i,m})$, maximizing expected counterfactual risk reduction over \textsc{Drop} with common-random-number channel draws to reduce variance under positive fading correlation, while Smooth-$L_1$ calibration anchors Softplus gains $u_{\theta,i}(a_m)$ to a positive Huber-location cardinal scale compatible with Eq.~\eqref{eq:convexhull} at inference. Alternating Eqs.~\eqref{eq:rxloss} and~\eqref{eq:txloss} runs $2{,}000$ joint steps on $126{,}982$ ImageNet images (batch $64$, AdamW $\text{lr} = 3\times 10^{-4}$, cosine decay, clip $1.0$).

\section{Experiments}
\label{sec:exp}

We benchmark AdapToC against separated codecs, continuous JSCC, static UEP, and iterative token decoders across channels, symbol budgets, tokenizer topologies, downstream tasks, and hardware latency (Tabs.~\ref{tab:main}, \ref{tab:generality}; Figs.~\ref{fig:curves} to~\ref{fig:latency}).

\begin{table*}[t]
\centering
\setlength{\abovecaptionskip}{1pt}
\setlength{\belowcaptionskip}{0.5pt}
\setlength{\aboverulesep}{0.35ex}
\setlength{\belowrulesep}{0.35ex}
\caption{\textbf{AdapToC leads perceptual and semantic resilience at $8\,\text{dB}$ and across $0$ to $10\,\text{dB}$ fading.}
All token methods share VQGAN-$f_{16}$ ($B\le 5{,}632$; $T{=}1$ AdapToC: $3.52\,\text{M}$ RX $+$ $8.16\,\text{M}$ TX, $0.67\,\text{ms}$ RX/$5.63\,\text{ms}$ E2E vs.\ $T{=}8$ TokenCom); \textbf{(a)} $8\,\text{dB}$ ($6/6$ fading wins), \textbf{(b)} $0$ to $10\,\text{dB}$ mean ($9/9$), gray: oracle, \textbf{bold}/\underline{underlined}: \#1/\#2.}
\label{tab:main}
\begin{subtable}[t]{\textwidth}
\centering
\renewcommand{\arraystretch}{0.86}
\caption{Paired $8\,\text{dB}$ Channel-Resilience Benchmark ($N=1{,}024$ ImageNet, $B\le 5{,}632$ symbols).}
\scriptsize
\setlength{\tabcolsep}{2.0pt}
\begin{tabular*}{\textwidth}{@{\extracolsep{\fill}}l rrrr rrrr rrrr@{}}
\toprule
& \multicolumn{4}{c}{\textbf{AWGN}} & \multicolumn{4}{c}{\textbf{Rician-$K6$}} & \multicolumn{4}{c}{\textbf{Rayleigh NLoS}} \\
\cmidrule(lr){2-5}\cmidrule(lr){6-9}\cmidrule(lr){10-13}
\textbf{Method} & LPIPS$\downarrow$ & rFID$\downarrow$ & CLIP@1$\uparrow$ & PSNR$\uparrow$ & LPIPS$\downarrow$ & rFID$\downarrow$ & CLIP@1$\uparrow$ & PSNR$\uparrow$ & LPIPS$\downarrow$ & rFID$\downarrow$ & CLIP@1$\uparrow$ & PSNR$\uparrow$ \\
\midrule
\multicolumn{13}{@{}l@{}}{\textcolor{bpGray}{\textit{Reference Oracles} (Noiseless VQGAN-$f_{16}$ $N=1{,}024$: LPIPS 0.1599, rFID 18.20, PSNR 21.21\,dB)}} \\
\textcolor{bpGray}{BPG + Shannon Cap.} & \textcolor{bpGray}{0.3935} & \textcolor{bpGray}{50.52} & \textcolor{bpGray}{47.07} & \textcolor{bpGray}{26.05} & \textcolor{bpGray}{0.4076} & \textcolor{bpGray}{52.38} & \textcolor{bpGray}{46.48} & \textcolor{bpGray}{25.84} & \textcolor{bpGray}{0.4243} & \textcolor{bpGray}{56.10} & \textcolor{bpGray}{44.92} & \textcolor{bpGray}{25.58} \\
\midrule
\multicolumn{13}{@{}l@{}}{\textit{Link-Adapted Digital \& Continuous JSCC}} \\
JPEG + AMC-LDPC & 0.7334 & 125.67 & 18.95 & 21.31 & 0.7848 & 158.42 & 10.35 & 20.43 & 0.8405 & 223.21 & 1.86 & 18.13 \\
WebP + AMC-LDPC & 0.6380 & 107.40 & 26.86 & \underline{22.38} & 0.6909 & 127.32 & 22.56 & 21.63 & 0.7681 & 162.08 & 12.40 & 19.91 \\
BPG + AMC-LDPC & 0.5550 & 82.19 & 38.09 & \textbf{23.76} & 0.6067 & 95.50 & \underline{32.52} & \textbf{22.92} & 0.6863 & 117.55 & 20.12 & 21.34 \\
DeepJSCC-Q (16QAM) & 0.6963 & 103.12 & 26.07 & 21.92 & 0.7116 & 109.30 & 24.71 & 21.74 & 0.7302 & 116.70 & 20.80 & \underline{21.47} \\
MambaJSCC-16QAM & 0.7016 & 98.80 & 28.32 & 22.05 & 0.7148 & 104.51 & 24.61 & \underline{21.87} & 0.7307 & \underline{110.28} & \underline{22.17} & \textbf{21.59} \\
\midrule
\multicolumn{13}{@{}l@{}}{\textit{Hard-Erasure Token Baselines (VQGAN-$f_{16}$; TokenCom $T=8$, TONIC Static UEP)}} \\
TokenCom (VQGAN, Fixed) & 0.4237 & 93.90 & 16.41 & 15.46 & 0.5223 & 119.37 & 8.69 & 13.97 & 0.5813 & 126.99 & 4.39 & 12.98 \\
TokenCom (VQGAN+AMC) & \textbf{0.1599} & \textbf{18.19} & \underline{54.59} & 21.21 & 0.4879 & 118.21 & 8.59 & 14.44 & 0.5419 & 130.03 & 5.76 & 13.61 \\
TONIC-Style Static UEP & 0.2675 & 41.09 & 44.34 & 18.62 & \underline{0.3537} & \underline{68.24} & 29.88 & 17.35 & \underline{0.4706} & 113.46 & 12.79 & 15.91 \\
\midrule
\multicolumn{13}{@{}l@{}}{\textit{Single-Pass Token Transceivers (VQGAN-$f_{16}$, $T=1$; AdapToC: $3.52\,\text{M}$ RX $+$ $8.16\,\text{M}$ TX, $0.67\,\text{ms}$ RX)}} \\
\textcolor{bpOurs}{\textbf{AdapToC (Ours)}} & \textbf{0.1599} & \underline{18.20} & \textbf{54.79} & 21.21 & \textbf{0.1805} & \textbf{21.63} & \textbf{53.22} & 20.67 & \textbf{0.2287} & \textbf{31.77} & \textbf{47.46} & 19.74 \\
\bottomrule
\end{tabular*}
\end{subtable}
\par\vspace{2pt}
\begin{subtable}[t]{\textwidth}
\centering
\renewcommand{\arraystretch}{0.86}
\caption{Multi-SNR Mean ($0$ to $10\,\text{dB}$, $N=4{,}974$, VQGAN-$f_{16}$). Noiseless: 0.1620 LPIPS / 5.36 rFID / 55.47\% CLIP@1 / 21.03\,dB.}
\scriptsize
\setlength{\tabcolsep}{2.0pt}
\begin{tabular*}{\textwidth}{@{\extracolsep{\fill}}l rrrr rrrr rrrr@{}}
\toprule
& \multicolumn{4}{c}{\textbf{AWGN ($0$ to $10\,\text{dB}$ Mean)}} & \multicolumn{4}{c}{\textbf{Rician-$K6$ ($0$ to $10\,\text{dB}$ Mean)}} & \multicolumn{4}{c}{\textbf{Rayleigh NLoS ($0$ to $10\,\text{dB}$ Mean)}} \\
\cmidrule(lr){2-5}\cmidrule(lr){6-9}\cmidrule(lr){10-13}
\textbf{Method} & LPIPS$\downarrow$ & rFID$\downarrow$ & CLIP@1$\uparrow$ & PSNR$\uparrow$ & LPIPS$\downarrow$ & rFID$\downarrow$ & CLIP@1$\uparrow$ & PSNR$\uparrow$ & LPIPS$\downarrow$ & rFID$\downarrow$ & CLIP@1$\uparrow$ & PSNR$\uparrow$ \\
\midrule
\textcolor{bpGray}{BPG + Shannon Cap.} & \textcolor{bpGray}{0.4635} & \textcolor{bpGray}{38.40} & \textcolor{bpGray}{42.55} & \textcolor{bpGray}{24.89} & \textcolor{bpGray}{0.4761} & \textcolor{bpGray}{40.31} & \textcolor{bpGray}{41.60} & \textcolor{bpGray}{24.71} & \textcolor{bpGray}{0.4936} & \textcolor{bpGray}{43.19} & \textcolor{bpGray}{40.36} & \textcolor{bpGray}{24.46} \\
\midrule
\multicolumn{13}{@{}l@{}}{\textit{Continuous JSCC Baselines}} \\
DeepJSCC-Q (16QAM) & 0.7261 & 87.15 & 19.44 & \underline{21.26} & 0.7368 & 91.63 & 17.45 & \underline{21.07} & 0.7497 & 97.88 & 14.87 & \underline{20.80} \\
MambaJSCC-16QAM & 0.7287 & 80.56 & 19.98 & \textbf{21.40} & 0.7373 & \underline{84.72} & \underline{18.16} & \textbf{21.21} & 0.7470 & \underline{90.10} & \underline{15.91} & \textbf{20.94} \\
\midrule
\multicolumn{13}{@{}l@{}}{\textit{Same-Tokenizer Visual Token Systems (VQGAN-$f_{16}$)}} \\
TokenCom (VQGAN, Fixed) & 0.7127 & 193.46 & 11.95 & 12.95 & 0.7244 & 158.82 & 6.15 & 11.89 & 0.7350 & 135.56 & 3.09 & 11.21 \\
TokenCom (VQGAN+AMC) & \underline{0.3797} & \underline{50.93} & \underline{28.00} & 16.61 & 0.5641 & 85.16 & 9.33 & 13.18 & 0.6551 & 99.61 & 2.51 & 11.69 \\
TONIC-Style Static UEP & 0.4564 & 85.43 & 25.48 & 16.49 & \underline{0.5630} & 114.12 & 13.92 & 15.04 & \underline{0.6346} & 136.74 & 6.10 & 14.24 \\
\textcolor{bpOurs}{\textbf{AdapToC (Ours)}} & \textbf{0.2516} & \textbf{26.57} & \textbf{41.86} & 19.69 & \textbf{0.2817} & \textbf{29.20} & \textbf{38.30} & 19.12 & \textbf{0.3312} & \textbf{38.42} & \textbf{31.27} & 18.42 \\
\bottomrule
\end{tabular*}
\end{subtable}
\end{table*}

\subsection{Experimental Setup}
\label{sec:setup}

\paragraph{Datasets and protocol.}
At $8\,\text{dB}$, $1{,}024$ ImageNet validation images ($384^2$, $L=576$, $B=5{,}632$, $1.02\,\text{b/sym}$ budget rate; $1.03\,\text{b/sym}$ over $5{,}568$ used symbols) span AWGN, Rician-$K6$, and Rayleigh fading (Tab.~\ref{tab:main}a: pilot $\text{NMSE}=0.02$, $0.2287$ LPIPS; Tab.~\ref{tab:generality}: clean-pilot Rayleigh $\text{NMSE}=0.0$, $0.2296$ LPIPS). Inside $[B_{\text{full}},B_{\text{up}})$, fading AdapToC enters zero-header ($0\text{h}$) mode ($36$ \textsc{Strong} anchors $+$ $540$ \textsc{Weak} tokens), so Tabs.~\ref{tab:main} and~\ref{tab:generality} (VQGAN) isolate soft recovery and $R_\psi$, while $384\text{h}$ allocation via $S_\theta$ governs VQGAN deficit/surplus budgets (Sec.~\ref{sec:ablations}); TiTok uses an $\text{AL}=1$ ($192$-sym) header and VAR piggybacks its schedule inside anchors ($\text{AL}=0$; Sec.~\ref{sec:limitations}). Multi-SNR and budget sweeps use $4{,}974$ leakage-free ImageNet images; multiple-choice knapsack (MCKP) policy audits use $512$; component ablations use $1{,}024$; zero-shot driving transfer uses $1{,}024$ NuPlan frames~\cite{caesar2020nuscenes,caesar2021nuplan,drivingworld2024}; and ImageNet-$100$ classification uses $5{,}000$ images ($100$ classes).

\paragraph{Metrics and baselines.}
We report AlexNet Learned Perceptual Image Patch Similarity (LPIPS, $\downarrow$)~\cite{zhang2018lpips}, reconstruction Fr\'echet Inception Distance (rFID, $\downarrow$), Structural Similarity Index (SSIM, $\uparrow$), CLIP ViT-B/32 top-1 accuracy and cosine similarity ($\uparrow$), and Peak Signal-to-Noise Ratio (PSNR, $\uparrow$) alongside paired $1{,}000$-bootstrap $95\%$ confidence intervals (CIs). Baselines comprise JPEG, WebP, and BPG~\cite{bellard2014bpg} (AMC/fixed $16$-QAM $R=1/2$, Shannon capacity oracle), DeepJSCC-Q~\cite{tung2022deepjsccq}, MambaJSCC~\cite{wu2025mambajscc} ($126{,}982$-image split), TONIC UEP~\cite{liu2026tonic}, and retrained $8$-step TokenCom~\cite{qiao2025tokencom} ($39.95\,\text{M}$ params on shared $K=1{,}024$ VQGAN-$f_{16}$, fixed and link-adapted AMC). Latent task heads train on noiseless latents ($4{,}800$ steps, CLIP distilled on $24{,}576$ images); pixel-decoded ViT-B/32 runs zero-shot.

\subsection{Main Results}
\label{sec:main_results}

\begin{figure*}[t]
  \centering
  \setlength{\abovecaptionskip}{2pt}
  \setlength{\belowcaptionskip}{1pt}
  \includegraphics[width=0.93\linewidth]{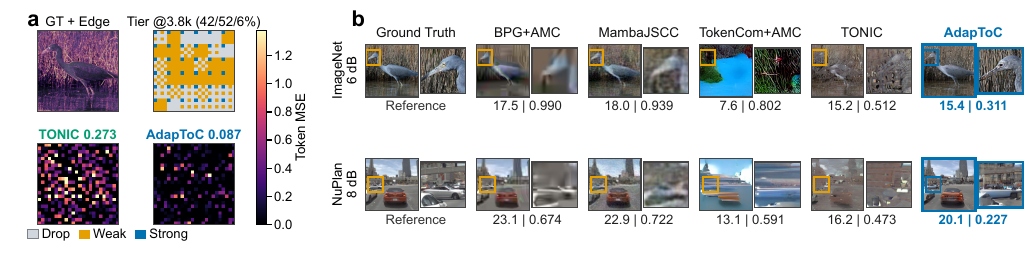}
  \caption{\textbf{AdapToC protects textured contours under tight budgets and cuts latent error.} (\textbf{a}, $\#6581$, $6\,\text{dB}$) $B{=}3{,}840$ drops $42\%$ background ($52\%$ weak, $6\%$ strong); $0\text{h}$ $B{=}5{,}632$ cuts MSE $0.273{\to}0.087$. (\textbf{b}) [Full\,$\mid$\,Zoom] renders (PSNR\,$\mid$\,LPIPS) on ImageNet $\#6581$ ($6\,\text{dB}$) and NuPlan $\#367$ ($8\,\text{dB}$).}
  \label{fig:qual}
\end{figure*}

\paragraph{Fading channel resilience.}
AdapToC leads all $6/6$ fading perceptual and semantic columns at $8\,\text{dB}$ (Tab.~\ref{tab:main}a) and all $9/9$ multi-SNR columns across $0$ to $10\,\text{dB}$ (Tab.~\ref{tab:main}b). At $8\,\text{dB}$ AWGN, AdapToC and $100\%$-CRC TokenCom match noiseless VQGAN ($0.1599$ LPIPS, $21.21\,\text{dB}$; $0.01$ rFID variance within decode precision). Against TONIC and TokenCom+AMC at $8\,\text{dB}$, AdapToC lowers LPIPS by $49.0\%$ and $63.0\%$ on Rician-$K6$ ($0.1805$ vs.\ $0.3537$ and $0.4879$) and by $51.4\%$ and $57.8\%$ on Rayleigh ($0.2287$ vs.\ $0.4706$ and $0.5419$; Fig.~\ref{fig:qual}b) while cutting Rayleigh rFID over $3.5\times$ ($31.77$ vs.\ $113.46$/$130.03$) with $+34.67$/$+41.70\,\text{pp}$ CLIP@1. Across $0$ to $10\,\text{dB}$ ($4{,}974$ images), Rayleigh LPIPS drops $47.8\%$ and $49.4\%$ ($0.3312$ vs.\ $0.6346$ and $0.6551$) and Rician-$K6$ drops $50.0\%$/$50.1\%$. Continuous MSE-trained JSCC and BPG exceed VQGAN's $21.21/21.03\,\text{dB}$ PSNR ceiling yet incur severe perceptual blur ($0.555$ to $0.750$ LPIPS).

\begin{figure*}[t]
  \centering
  \setlength{\abovecaptionskip}{2pt}
  \setlength{\belowcaptionskip}{1pt}
  \includegraphics[width=0.92\linewidth]{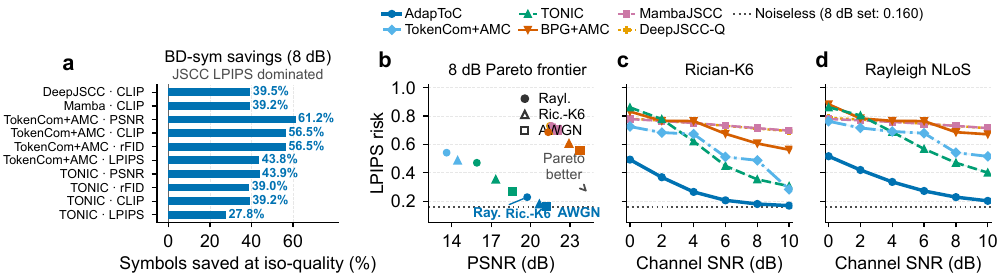}
  \caption{\textbf{AdapToC saves symbols over digital and UEP baselines and avoids Rayleigh cliffs.} (\textbf{a}) $8\,\text{dB}$ Rician-$K6$ BD-symbol savings vs.\ TokenCom+AMC ($43.8\%$ to $61.2\%$) and TONIC ($27.8\%$ to $43.9\%$). (\textbf{b}) $8\,\text{dB}$ R-D tradeoffs. (\textbf{c}, \textbf{d}) Fading sweeps.}
  \label{fig:curves}
\end{figure*}

\paragraph{Budget efficiency and SNR sweeps.}
Across budgets $B \in \{2{,}816,\allowbreak 4{,}224,\allowbreak 5{,}632,\allowbreak 8{,}448\}$ at $8\,\text{dB}$ Rician-$K6$ ($N=4{,}974$; Fig.~\ref{fig:curves}a), AdapToC saves $43.8\%$ LPIPS, $56.5\%$ rFID, $56.5\%$ CLIP@1, $61.2\%$ PSNR, and $61.7\%$ SSIM Bj{\o}ntegaard-Delta (BD) symbols over TokenCom+AMC ($27.8\%$ to $57.4\%$ vs.\ TONIC; $39.2\%$/$39.5\%$ CLIP@1 and $58.6\%$/$63.0\%$ rFID vs.\ MambaJSCC/DeepJSCC-Q, dominating both in LPIPS across all budgets). AdapToC likewise leads Rayleigh LPIPS across every SNR from $0$ to $10\,\text{dB}$ (Fig.~\ref{fig:curves}c,d).

\begin{table*}[t]
\centering
\setlength{\abovecaptionskip}{1pt}
\setlength{\belowcaptionskip}{0.5pt}
\setlength{\belowrulesep}{0.38ex}
\renewcommand{\arraystretch}{0.86}
\caption{\textbf{AdapToC transfers across 1D, multi-scale, and 2D tokenizers on ImageNet and zero-shot NuPlan.} At $8\,\text{dB}$ Rayleigh ($N=1{,}024$, with AMC BPG), AdapToC-PnP/Joint take the top two perceptual/semantic ranks on every tokenizer; \textbf{bold}/\underline{underlined}: \#1/\#2.}
\label{tab:generality}
\scriptsize
\setlength{\tabcolsep}{2.8pt}
\begin{tabular*}{\textwidth}{@{\extracolsep{\fill}}l rrrr rrrr@{}}
\toprule
& \multicolumn{4}{c}{\textbf{ImageNet ($N=1{,}024$)}} & \multicolumn{4}{c}{\textbf{NuPlan ($N=1{,}024$)}} \\
\cmidrule(lr){2-5}\cmidrule(lr){6-9}
\textbf{Tokenizer / Method} & LPIPS$\downarrow$ & rFID$\downarrow$ & Cos$\uparrow$ & PSNR$\uparrow$ & LPIPS$\downarrow$ & rFID$\downarrow$ & Cos$\uparrow$ & PSNR$\uparrow$ \\
\midrule
\multicolumn{9}{@{}l@{}}{\textbf{1D Global Sequence: TiTok-L32 ($B=512$, $32$ tokens)} \textcolor{bpGray}{\textit{(Noiseless: IN 0.3316/28.05/15.88; NuPlan 0.3323/56.83/16.99\,dB)}}} \\
\quad BPG + AMC-LDPC & 0.9190 & 211.99 & 0.511 & 14.77 & 0.9185 & 334.96 & 0.522 & \underline{17.79} \\
\quad MambaJSCC-16QAM & 0.7448 & 147.20 & 0.554 & \textbf{18.65} & 0.7656 & 295.38 & 0.577 & \textbf{19.67} \\
\quad TONIC-UEP & 0.5341 & 54.13 & 0.725 & 13.36 & 0.5304 & 112.39 & 0.688 & 15.06 \\
\quad \textcolor{bpOurs}{\textbf{AdapToC-PnP (Ours, 0-Param)}} & \textbf{0.4293} & \textbf{38.35} & \textbf{0.803} & 14.75 & \textbf{0.4105} & \textbf{65.71} & \textbf{0.760} & 15.91 \\
\quad \textcolor{bpOurs}{\textbf{AdapToC-Joint (Ours, Co-Trained)}} & \underline{0.4394} & \underline{39.75} & \underline{0.794} & \underline{14.80} & \underline{0.4222} & \underline{68.06} & \underline{0.753} & 15.81 \\
\midrule
\multicolumn{9}{@{}l@{}}{\textbf{Multi-Scale Pyramid: VAR-VAE ($B=4{,}608$, $10$ scales)} \textcolor{bpGray}{\textit{(Noiseless: IN 0.1078/12.87/22.10; NuPlan 0.1117/16.03/23.23\,dB)}}} \\
\quad BPG + AMC-LDPC & 0.6708 & 121.89 & 0.633 & \underline{20.93} & 0.6898 & 229.45 & 0.676 & \underline{22.11} \\
\quad MambaJSCC-16QAM & 0.6406 & 114.25 & 0.672 & \textbf{21.47} & 0.6646 & 250.64 & 0.697 & \textbf{22.88} \\
\quad TONIC-UEP & 0.4718 & 111.94 & 0.671 & 15.61 & 0.4500 & 156.79 & 0.697 & 16.90 \\
\quad \textcolor{bpOurs}{\textbf{AdapToC-PnP (Ours, 0-Param)}} & \underline{0.4034} & \underline{92.26} & \underline{0.720} & 16.69 & \textbf{0.3950} & \underline{134.82} & \underline{0.739} & 17.88 \\
\quad \textcolor{bpOurs}{\textbf{AdapToC-Joint (Ours, Co-Trained)}} & \textbf{0.3986} & \textbf{89.93} & \textbf{0.735} & 16.61 & \underline{0.4040} & \textbf{127.32} & \textbf{0.741} & 17.69 \\
\midrule
\multicolumn{9}{@{}l@{}}{\textbf{2D Spatial Grid: VQGAN-$f_{16}$ ($B=5{,}632$, $24\times 24$)} \textcolor{bpGray}{\textit{(Noiseless: IN 0.1599/18.20/21.21; NuPlan 0.1631/28.65/21.85\,dB)}}} \\
\quad BPG + AMC-LDPC & 0.6887 & 118.00 & 0.627 & \underline{21.28} & 0.6775 & 222.31 & 0.659 & \underline{22.49} \\
\quad MambaJSCC-16QAM & 0.7308 & 111.10 & 0.664 & \textbf{21.58} & 0.7611 & 244.81 & 0.687 & \textbf{22.69} \\
\quad TokenCom (Fixed) & 0.5847 & 126.71 & 0.603 & 12.94 & 0.5946 & 218.19 & 0.559 & 13.45 \\
\quad TokenCom + Adaptive MCS & 0.5421 & 131.84 & 0.615 & 13.57 & 0.5479 & 208.95 & 0.563 & 14.28 \\
\quad TONIC-UEP & 0.4756 & 114.37 & 0.664 & 15.85 & 0.4207 & 211.82 & 0.609 & 17.25 \\
\quad \textcolor{bpOurs}{\textbf{AdapToC-PnP (Ours, 0-Param)}} & \underline{0.2643} & \underline{40.02} & \underline{0.851} & 18.91 & \underline{0.2605} & \underline{66.13} & \underline{0.819} & 19.92 \\
\quad \textcolor{bpOurs}{\textbf{AdapToC-Joint (Ours, Co-Trained)}} & \textbf{0.2296} & \textbf{31.42} & \textbf{0.882} & 19.75 & \textbf{0.2300} & \textbf{53.66} & \textbf{0.859} & 20.47 \\
\bottomrule
\end{tabular*}
\end{table*}

\subsection{Ablation and Diagnostic Analysis}
\label{sec:ablations}

\begin{figure}[!t]
  \centering
  \setlength{\abovecaptionskip}{2pt}
  \setlength{\belowcaptionskip}{1pt}
  \includegraphics[width=0.91\linewidth]{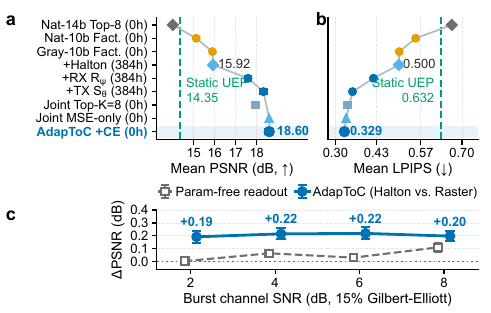}
  \caption{\textbf{Alignment, soft recovery, CE loss, and Halton compound.} (\textbf{a}, \textbf{b}) Rayleigh $0$ to $10\,\text{dB}$ ($95\%$ CIs) show compaction, Pseudo-Gray, $R_\psi$, $S_\theta$, $0\text{h}$/$384\text{h}$, and $+\mathcal{L}_{\text{CE}}$. (\textbf{c}) Burst $\Delta\text{PSNR}$.}
  \label{fig:ablation}
\end{figure}

\paragraph{Component and loss ablations.}
Every stage compounds monotonically over unaligned readout (Fig.~\ref{fig:ablation}a,b, tagging $0\text{h}$ vs.\ $384\text{h}$). Across $0$ to $10\,\text{dB}$ Rayleigh fading ($N=1{,}024$), $14$-bit Top-$8$ readout ($0\text{h}$, $14.01\,\text{dB}$, $0.6678$ LPIPS) trails static UEP ($14.35\,\text{dB}$, $0.6324$), whereas $10$-bit compaction with exact readout reaches $15.12\,\text{dB}$ ($0.5536$) and Pseudo-Gray alignment ($-61.1\%$ $d_H=1$ MSE) reaches $15.89\,\text{dB}$ ($0.5019$). Entering $384\text{h}$ mode to add Halton dispersion ($15.92$), $R_\psi$ ($17.59$), and $S_\theta$ ($18.32\,\text{dB}$, $0.3543$), followed by joint co-training with $0\text{h}$ switching, reaches $18.60\,\text{dB}$ and $0.3286$ LPIPS ($64.48$ rFID; $0.3366$ Raster) and beats matched $10$-bit Top-$8$ readout ($1.86$ vs.\ $0.041\,\text{ms}$, $17.96\,\text{dB}$, $0.3372$). In a controlled $2{,}000$-step benchmark ($N=1{,}024$), $\lambda_{\text{ce}}=0.02$ in Eqs.~\eqref{eq:rxloss} and~\eqref{eq:txloss} improves $0\text{h}$ LPIPS ($0.3338 \to 0.3292$), rFID ($66.62 \to 64.67$), CLIP cosine ($0.7841 \to 0.7871$), and Sobel edge retention ($78.88\% \to 79.88\%$) at constant $18.59\,\text{dB}$ PSNR ($0.3549 \to 0.3495$ in $384\text{h}$); at $0$ and $8\,\text{dB}$, it yields paired $\Delta\text{LPIPS} = -0.0110$ $[-0.0115,\allowbreak -0.0105]$ and $-0.0014$ $[-0.0016,\allowbreak -0.0013]$, $+1.60\%$ $[+1.46,\allowbreak +1.74]$ and $+0.67\%$ $[+0.62,\allowbreak +0.74]$ edge gain, and rFID $138.65 \to 132.69$ and $31.30 \to 31.00$.

\paragraph{Budget regimes, tier granularity, and mixed-policy utility.}
Across Rayleigh budgets ($N=512$; Fig.~\ref{fig:teaser}d), comparing AdapToC against static \textsc{Weak} ($0\text{h}$) and variance $L_2$-MCKP ($\|z_i^q-\bar{z}^q\|_2^2$, $384\text{h}$) empirically validates the overhead trade-off of Corollary~\ref{cor:granularity}. In severe deficit ($B \le 4{,}224$; Fig.~\ref{fig:qual}a), $S_\theta$ beats static \textsc{Weak} ($\Delta\text{LPIPS} = -0.0153\,[-0.0195,-0.0110]$ at $2{,}048$ and $-0.0244\,[-0.0293,-0.0189]$ at $2{,}816$, $4\,\text{dB}$) and $L_2$-MCKP ($-0.0136$ at $3{,}840$), while at $B=5{,}184 = B_{\text{full}}-B_{\text{hdr}}$ header truncation ($0.3856$ LPIPS) empirically edges $384\text{h}$ ($0.3977$) in LPIPS (whereas $384\text{h}$ leads PSNR by $+0.26\,\text{dB}$), illustrating the overhead crossover region. Inside $[B_{\text{full}},B_{\text{up}})$ ($B \in \{5{,}568, 5{,}632, 5{,}856\}$), $0\text{h}$ matches static ($0.3400/0.3402/0.3367$ vs.\ $384\text{h}$ $0.3634/0.3562/0.3427$ at $4\,\text{dB}$; $0.2328/0.2348/0.2339$ vs.\ $0.2583/0.2514/0.2398$ at $8\,\text{dB}$). Above $B_{\text{up}}$, static $0\text{h}$ saturates ($0.3400/0.2342$ at $8{,}448$) whereas $S_\theta$ promotes anchors to reach $0.3017$ ($\Delta = -0.0383\,[-0.0423,-0.0344]$ at $4\,\text{dB}$; $0.2202$ at $8\,\text{dB}$). Under matched $384\text{h}$ MCKP allocation, adding the optional fourth $16$-QAM \textsc{Weak} rung empirically ties three tiers at $7{,}680$ ($0.3025/0.2173$ vs.\ $0.3026/0.2171$ at $4/8\,\text{dB}$; two-tier $0.4131/0.3604$), while $0\text{h}$ ($0.3434/0.2359$) leads at $5{,}632$. Across $4{,}608$ whole-image mixed-policy perturbations ($128$ images $\times 36$ units with full $R_\psi$ coupling, $N=128$ cluster $2{,}000$-bootstrap CIs, permutation tie-breaking matching average-rank $\Delta\rho_s$ to three decimals, and ratio-of-sums excess top-quartile gain capture $\sum(\text{top-}Q_1-0.25\,\text{tot})/\sum(\text{ora-}Q_1-0.25\,\text{tot})$ with zero-denominator guard), co-trained $S_\theta$ leads both \textsc{Drop}$\to\allowbreak$\textsc{Weak} ($3{,}840$, $4\,\text{dB}$: $\rho_s = 0.404/0.335$ MSE/LPIPS vs.\ $L_2$ $0.390/0.314$, $\Delta\rho_s = +0.022\,[+0.009,+0.035]$, $33.0\%$ vs.\ $30.1\%$) and \textsc{Weak}$\to\allowbreak$\textsc{Strong} ($7{,}680$, $4\,\text{dB}$: $\Delta\rho_s = +0.043\,[+0.021,+0.066]$, $15.2\%$ vs.\ $9.6\%$), while $8\,\text{dB}$ saturation shrinks mean gain to $0.0012$ ($\rho_s = 0.051/0.029$).

\paragraph{Soft recovery, bursts, and CSI noise.}
Isolating CRC-failed Rayleigh blocks ($0$ to $10\,\text{dB}$) drops latent MSE from $0.5550$ (omission) to $0.4459$ (soft readout) and $0.3508$ ($R_\psi$), yielding $36.8\%$ pooled ($37.8\%$ per-SNR mean) failed-subset improvement $\hat{\eta}$ over hard omission and $50.4\%$ at $8\,\text{dB}$ Rician-$K6$. At $0\text{h}$ $B=5{,}632$, soft LLRs preserve $100\%$ observation across $12.1\%$ to $80.3\%$ CRC pass rates (Fig.~\ref{fig:physim}c) and pull $0\,\text{dB}$ \textsc{Weak} $18.2\%$ below the soft chord ($0.384<0.470$; hard chord $0.552$; Fig.~\ref{fig:teaser}c), empirically illustrating Theorem~\ref{thm:unfolding}. Halton interleaving adds $+0.19$ to $+0.22\,\text{dB}$ under $15\%$ Gilbert--Elliott bursts (Fig.~\ref{fig:ablation}c), and calibrated pilot scaling sustains $0.3851$ LPIPS at $\text{NMSE}=0.30$ vs.\ $0.6445$ (TONIC) and $0.7163$ (TokenCom+AMC, Fig.~\ref{fig:physim}d).

\subsection{Cross-Tokenizer Transfer, Tasks, and Latency}
\label{sec:generality}

\paragraph{Cross-tokenizer transfer and 2D spatial bias.}
Zero-parameter AdapToC-PnP and co-trained AdapToC-Joint take the top two ranks in LPIPS, rFID, and CLIP cosine across TiTok-L32, VAR-VAE, and VQGAN-$f_{16}$ (Tab.~\ref{tab:generality}), confirming topology-agnostic transfer of factorized soft readout (Eq.~\eqref{eq:readout}). Comparing Joint against PnP reveals the physical 2D spatial inductive bias of our restoration pipeline: 2D Halton dispersion, anchor tiering, and ConvNeXt-attention restorer $R_\psi$ exploit local 2D neighborhood smoothness, cutting $8\,\text{dB}$ Rayleigh LPIPS a further $13.1\%$ on ImageNet ($0.2643 \to 0.2296$) and $11.7\%$ on NuPlan ($0.2605 \to 0.2300$) over PnP on 2D VQGAN-$f_{16}$ ($0.67\,\text{ms}$ RX) and strictly improving in-domain ImageNet metrics on multi-scale 2D VAR-VAE ($0.3986$ vs.\ $0.4034$ LPIPS; $89.93$ vs.\ $92.26$ rFID; $0.735$ vs.\ $0.720$ Cos, with NuPlan PnP edging LPIPS $0.3950$ vs.\ $0.4040$). Conversely, on compact 1D global TiTok-L32 ($32$ tokens without a physical 2D coordinate grid), lightweight latent-MSE residual regression nudges PSNR ($14.80$ vs.\ $14.75\,\text{dB}$) while slightly smoothing global latents relative to strictly convex-hull PnP ($0.4293$ vs.\ $0.4394$ LPIPS, $0.803$ Cos), making parameter-free PnP preferable for 1D global sequences.

\begin{figure}[!t]
  \centering
  \setlength{\abovecaptionskip}{2pt}
  \setlength{\belowcaptionskip}{1pt}
  \includegraphics[width=0.87\linewidth]{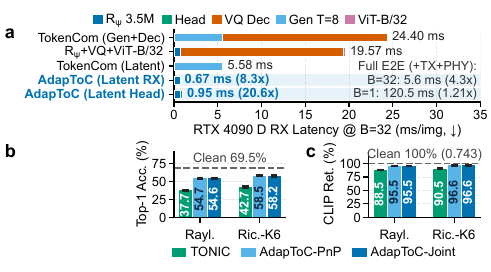}
  \caption{\textbf{Decoder-free inference boosts accuracy at $20.6\times$ lower RX latency ($8.3\times$ vs.\ TokenCom).} (\textbf{a}) RTX~4090\,D (batch $32$) latency. (\textbf{b}, \textbf{c}) $0$ to $10\,\text{dB}$ top-1 and CLIP retention ($95\%$ CIs).}
  \label{fig:latency}
\end{figure}

\paragraph{Decoder-free recognition and latency.}
Latent task heads reach $54.60\%/58.16\%$ Rayl./Ric.-$K6$ top-1 ($54.72\%/58.49\%$ PnP; $69.52\%$ clean), beating TONIC ($37.65\%/42.70\%$, $+16.95$/$+15.46\,\text{pp}$) and pixel-decoded MambaJSCC/BPG ($41.05\%/25.92\%$ Rayl.) while matching pixel-decoded ViT-B/32 ($54.48\%$, $N{=}1{,}000$) and retaining $0.709/0.718$ CLIP Cos ($95.5\%/96.6\%$; Fig.~\ref{fig:latency}b,c). On RTX~4090\,D (batch $32$; Fig.~\ref{fig:latency}a), $R_\psi{+}$head ($0.67{+}0.28\,\text{ms}$) runs in $0.95$ vs.\ $19.57\,\text{ms}$ pixel-decoded ($20.58\times$ RX; $3.64\times$ at batch $1$, $9.55$ vs.\ $34.75\,\text{ms}$) and $8.28\times$ faster than $T{=}8$ TokenCom ($5.58\,\text{ms}$); end-to-end latency including transmitter ($3.20\,\text{ms}$) and PHY LDPC ($1.49\,\text{ms}$) is $5.63$ vs.\ $24.25\,\text{ms}$ ($4.30\times$; $120.48$/$145.68\,\text{ms}$ at batch $1$).

\subsection{Limitations and Future Work}
\label{sec:limitations}
AdapToC-Joint matches module topologies across backbones---2D ConvNeXt-attention on VQGAN, a $2$-layer selector ($8$ orders) and $3$-layer Transformer on TiTok, and a $16\times 16$ ConvResNet, $2$-layer restorer, and $12$ coarse/$24$ fine groups on VAR---where spatial $R_\psi$ is natively tuned for 2D grids and frozen codebooks plateau near noiseless references ($21.21/15.88/22.10\,\text{dB}$). Designing unified topology-agnostic restoration (including generative objectives for 1D global sequences), temporal video priors~\cite{xie2026videorae}, and multi-round soft hybrid automatic repeat request (HARQ) remains future work.

\section{Conclusion}
\label{sec:conclusion}

We presented AdapToC, a single-pass transceiver that restores soft visual token posteriors across adaptive regimes using frozen codebooks and zero decoder gradients. Across $0$ to $10\,\text{dB}$ fading on VQGAN, TiTok, VAR, and zero-shot NuPlan, it improves perceptual and semantic fidelity over separated codecs, JSCC, static UEP, and TokenCom+AMC, saves $43.8\%$ to $61.7\%$ symbols, and raises accuracy while speeding RX recognition $20.6\times$ ($8.3\times$ vs.\ resampling).

{
    \small
    \bibliographystyle{ieeenat_fullname}
    \bibliography{main}
}

\end{document}